\documentclass{article}

\PassOptionsToPackage{numbers, compress}{natbib}

\usepackage[main,final]{neurips_2026}

\usepackage[utf8]{inputenc} % allow utf-8 input
\usepackage[T1]{fontenc}    % use 8-bit T1 fonts
\usepackage{hyperref}       % hyperlinks
\usepackage{url}            % simple URL typesetting
\usepackage{booktabs}       % professional-quality tables
\usepackage{amsfonts}       % blackboard math symbols
\usepackage{nicefrac}       % compact symbols for 1/2, etc.
\usepackage{microtype}      % microtypography
\usepackage{enumitem}
\usepackage{graphicx}
\usepackage{caption}
\usepackage{subcaption}
\usepackage{mathtools}
\usepackage{multirow}
\usepackage{float}
\usepackage[table]{xcolor}
\usepackage[ruled,vlined]{algorithm2e}

\title{Accelerating Rectified Flow Models \\ via Trajectory-Aware Caching}

\author{
  \mdseries
  \textbf{Xiao Liu}$^{1,2}$\thanks{Equal contribution.},\enspace
  \textbf{Kai Liu}$^{1}$\footnotemark[1],\enspace
  \textbf{Naiyang Guan}$^{2}$,\enspace
  \textbf{Hongliang Lu}$^{1}$,\enspace \\
  \textbf{Zhixin Wang}$^{3}$,\enspace
  \textbf{Zhikai Chen}$^{3}$,\enspace
  \textbf{Renjing Pei}$^{3}$,\enspace
  \textbf{Yulun Zhang}$^{1}$\thanks{Corresponding author: Yulun Zhang, yulun100@gmail.com} \\
  \textsuperscript{1}Shanghai Jiao Tong University,\enspace
  \textsuperscript{2}Academy of Military Science \\
  \textsuperscript{3}Huawei Noah's Ark Lab \\
  {\small\texttt{xiaoliu2000@sjtu.edu.cn}}
  \vspace{-5mm}
}

\begin{document}

\maketitle
\vspace{-2mm}
\begin{abstract}
\vspace{-2mm}
Diffusion and rectified flow (RF) models generate high-fidelity images and videos, but their iterative velocity-field evaluations are computationally expensive. Existing caching methods accelerate sampling by skipping timesteps, yet their coarse approximations introduce accumulated errors over long skip intervals and degrade quality under aggressive acceleration. We propose \textbf{TACache (Trajectory-Aware Cache)}, a training-free acceleration framework following a \textbf{skip-then-compensate} paradigm. TACache performs an orthogonal decomposition of discrete velocity acceleration along the RF trajectory into a parallel component and an orthogonal residual, isolating the magnitude and directional sources of per-step approximation error. The framework operates in two stages: \emph{offline}, cumulative variation thresholds on the magnitude and direction indicators yield the skip schedule and bound how far each skip interval may extend; \emph{online}, at each skipped step the offline statistics are combined with the sample's historical orthogonal direction to reconstruct the skipped velocity without additional model evaluations. Main comparisons on BAGEL, FLUX.1-dev, and Wan2.1-1.3B show that TACache achieves up to $4.11\times$ speedup on text-to-image generation and $2.11\times$ speedup on text-to-video generation, with consistent improvements over prior cache-based methods on most quality metrics. Our code is available at \url{https://github.com/XiaoLiu1237/TACache}.
\end{abstract}

\setlength{\abovedisplayskip}{2pt}
\vspace{-4mm}
\section{Introduction}
\vspace{-2mm}
\label{Introduction}
Recent advances in diffusion and rectified flow (RF) models have significantly improved the quality of image and video generation, but their iterative sampling procedure imposes a substantial computational burden: producing a single sample requires tens of expensive velocity-field evaluations along the sampling trajectory. The cost becomes prohibitive at scale, especially for high-resolution images and long videos. Generating a five-second clip with HunyuanVideo~\cite{kong2024hunyuanvideo} on an A100 GPU can take nearly an hour, severely limiting the deployment of these models in real applications.

\begin{figure}[t]
\centering
\vspace{-0.6em}
\includegraphics[width=\textwidth]{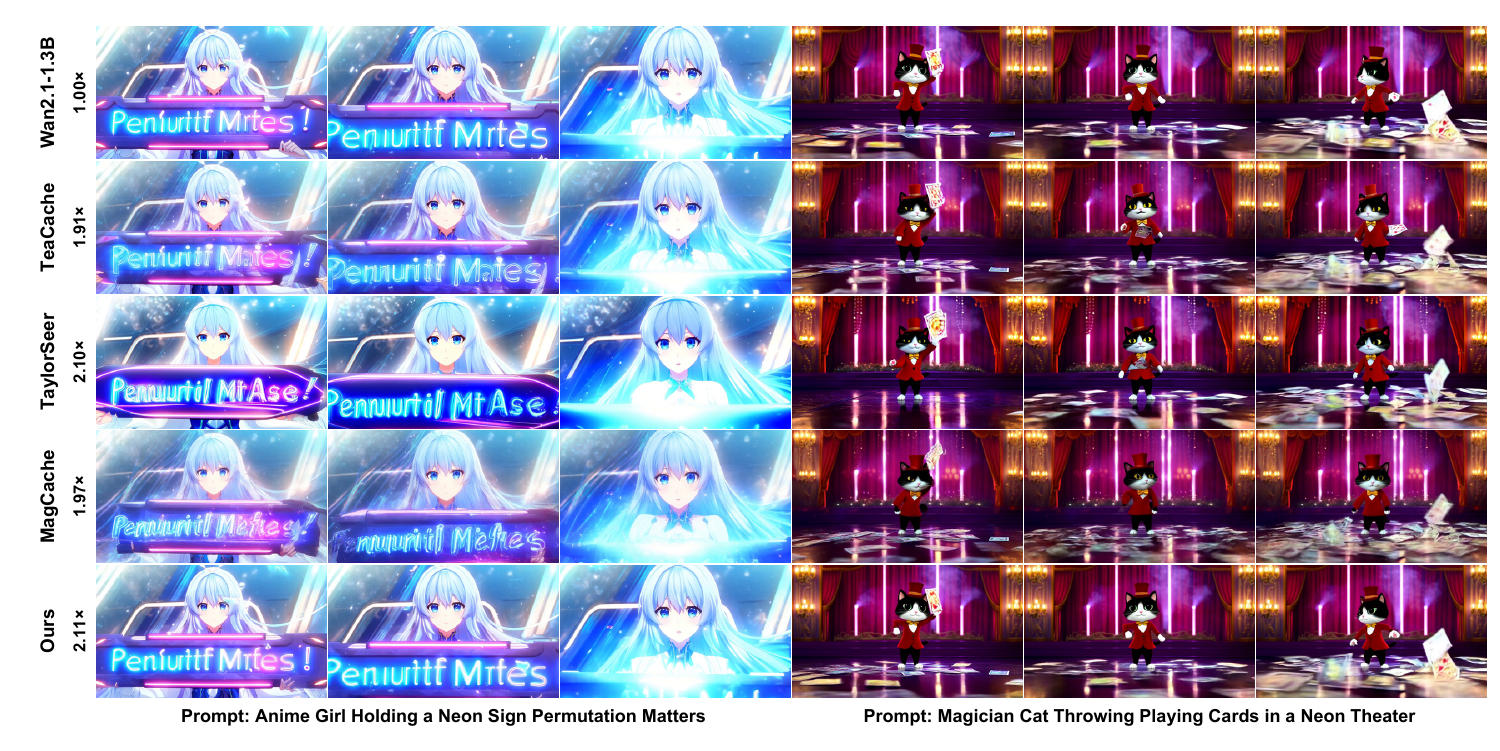}
\caption{Compared with other methods, TACache achieves a maximum speedup of 2.11$\times$ on Wan2.1-1.3B~\cite{wan} while preserving strong visual quality, demonstrating its effectiveness.}
\label{fig:show}
\vspace{-2.5em}
\end{figure}
Figure~\ref{fig:show} provides a qualitative overview of this speed-quality behavior on Wan2.1-1.3B.
A widely studied acceleration line is cache-based methods: rather than running a full model evaluation at every timestep, the model is evaluated only at key steps, and outputs at skipped steps are reconstructed from previous computations, with no additional model evaluation. Existing methods differ in approximation form: MagCache~\cite{magcache} models the scalar magnitude ratio of the input-relative residual; TeaCache~\cite{teacache} and EasyCache~\cite{easycache} directly reuse historical features or residuals; TaylorSeer~\cite{taylorseer} and DiCache~\cite{dicache} extrapolate via Taylor-style expansions. Despite these differences, all of these methods model the evolution of the velocity field at \textbf{a coarse granularity}, collapsing the trajectory dynamics into a single approximation pathway and leaving magnitude and directional variations without independently controllable mechanisms. As the skip interval grows, this coupled error accumulates rapidly and manifests as clear quality degradation under aggressive acceleration.

\begin{figure*}[t!]
  \centering
  \captionsetup[subfigure]{skip=0pt,font=footnotesize}
  % \captionsetup[subfigure]{font=footnotesize}
  \begin{subfigure}[t]{0.33\textwidth}
    \centering
    \includegraphics[width=\linewidth]{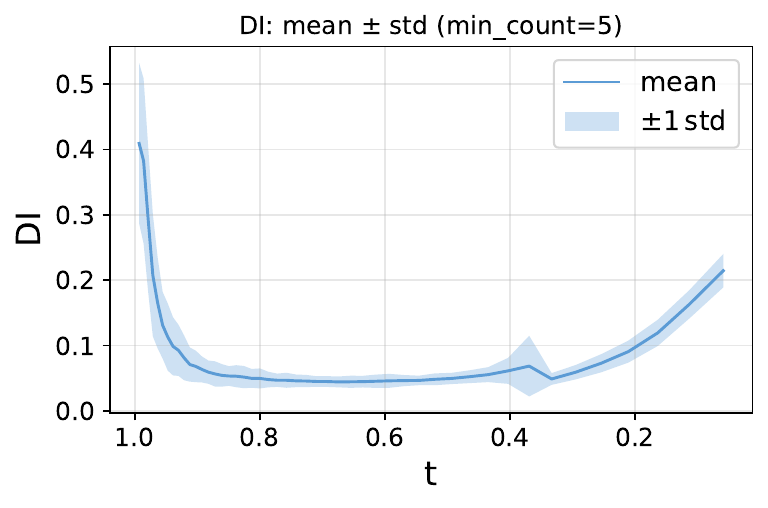}
    \caption{DI of BAGEL}
    \label{fig:r_bagel}
  \end{subfigure}\hfill
  \begin{subfigure}[t]{0.33\textwidth}
    \centering
    \includegraphics[width=\linewidth]{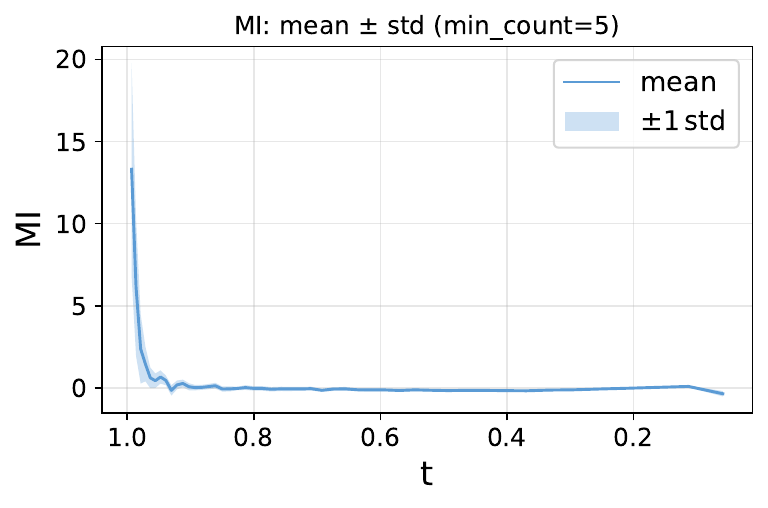}
    \caption{MI of BAGEL}
    \label{fig:k_bagel}
  \end{subfigure}\hfill
  \begin{subfigure}[t]{0.33\textwidth}
    \centering
    \includegraphics[width=\linewidth]{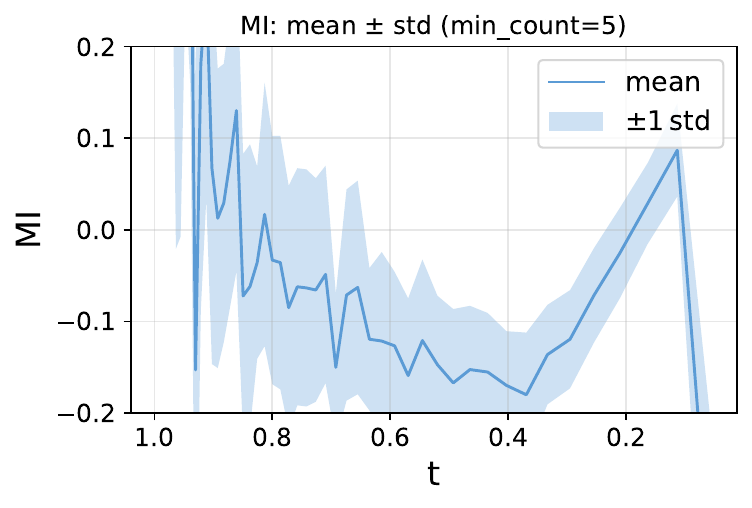}
    \caption{Zoomed-in MI curve on BAGEL}
    \label{fig:kzoom_bagel}
  \end{subfigure}
  \begin{subfigure}[t]{0.33\textwidth}
    \centering
    \includegraphics[width=\linewidth]{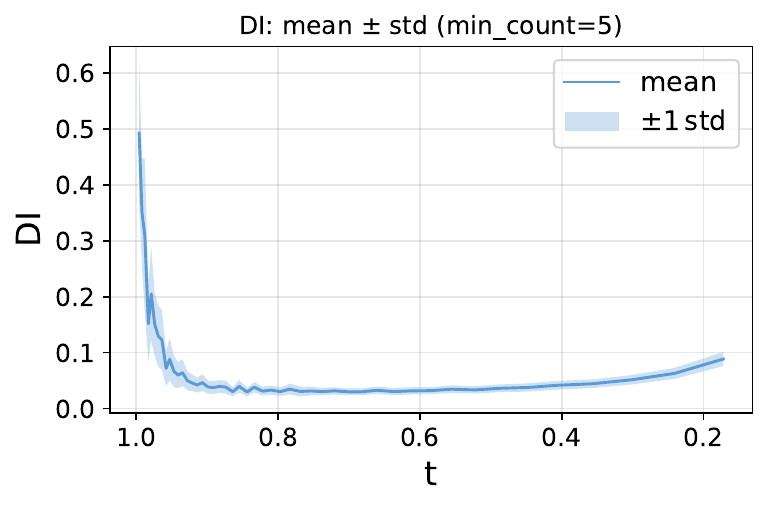}
    \caption{DI of Wan2.1-1.3B}
    \label{fig:r_wan}
  \end{subfigure}\hfill
  \begin{subfigure}[t]{0.33\textwidth}
    \centering
    \includegraphics[width=\linewidth]{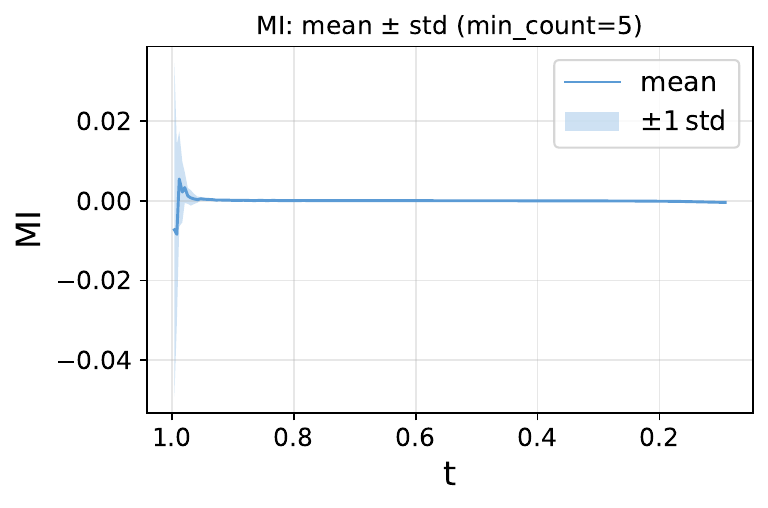}
    \caption{MI of Wan2.1-1.3B}
    \label{fig:k_wan}
  \end{subfigure}\hfill
  \begin{subfigure}[t]{0.33\textwidth}
    \centering
    \includegraphics[width=\linewidth]{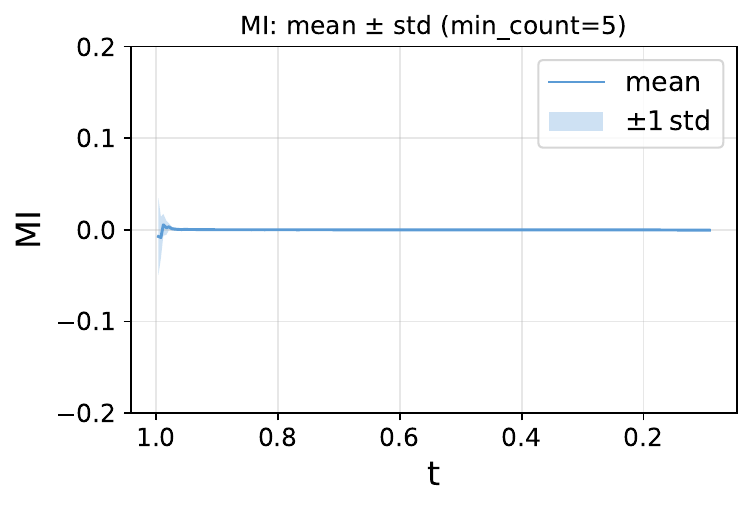}
    \caption{Zoomed-in MI curve on Wan2.1-1.3B}
    \label{fig:kzoom_wan}
  \end{subfigure}
  \vspace{-2mm}
  \caption{Evolution of the magnitude indicator (MI) and direction indicator (DI) over timesteps for different models. Both MI and DI remain relatively stable during intermediate timesteps, providing strong motivation and support for our step-skipping strategy.}
  \label{fig:MI-DI}
  \vspace{-4mm}
\end{figure*}

By decomposing the discrete acceleration of the RF velocity field into a parallel component capturing magnitude change and an orthogonal residual capturing directional drift (Fig.~\ref{fig:MI-DI}), we observe two complementary forms of stability across the text-to-image (T2I) and text-to-video (T2V) models we study. The per-timestep statistics of both terms are highly consistent across samples and evolve smoothly along the sampling trajectory. This dual stability indicates that the dynamics of the two terms admit a faithful offline characterization, providing a principled basis for replacing per-step online computation with stable offline statistics.

Building on this observation, we propose \textbf{TACache (Trajectory-Aware Cache)}, a training-free acceleration framework following a \textbf{skip-then-compensate} paradigm. TACache comprises three coordinated modules. \textbf{Parallel-Orthogonal Velocity Decomposition (POVD)} decomposes the discrete acceleration into a parallel component and an orthogonal residual, providing the geometric basis for both scheduling and reconstruction. For \emph{when to skip}, the \textbf{Stable Step Calculator (SSC)} constrains the cumulative variation of the magnitude and direction indicators within their respective thresholds, producing a safe skip schedule. For \emph{how to compensate}, the \textbf{Trajectory-Aware Skip-Update (TASU)} reconstructs skipped velocities by combining offline statistics with each sample's historical orthogonal direction, requiring no additional model evaluation. This shared parallel-orthogonal coordinate system allows TACache to control magnitude and direction errors separately, sustaining stable generation quality under high speedup.

We evaluate TACache on three representative RF generative models: BAGEL~\cite{bagel} and FLUX.1-dev~\cite{flux} for T2I, and Wan2.1-1.3B~\cite{wan} for T2V. In the main comparisons, TACache achieves up to $4.11\times$ speedup on text-to-image and $2.11\times$ speedup on text-to-video generation, with consistent improvements over prior cache-based methods on all reference-based fidelity metrics. On BAGEL, TACache also raises GenEval accuracy from $85.91\%$ to $87.32\%$ over the baseline.

\begin{itemize}[nosep,leftmargin=1.2em,itemindent=0pt]
  \item \textbf{A parallel-orthogonal error view for cached skipping.}
  We analyze cache-based skipping in rectified flow models through the velocity field rather than through generic feature reuse. By decomposing discrete velocity acceleration into a parallel magnitude term and an orthogonal residual, we derive a per-step error bound that separates skipping error into magnitude error, orthogonal strength error, and direction alignment error.

  \item \textbf{TACache: component-wise skip then compensate.}
  Built on this decomposition, we introduce TACache, a training-free framework that controls the three error terms with offline and online modules. POVD extracts population-level magnitude and direction indicators, SSC converts them into a skip schedule through cumulative-variation constraints, and TASU reconstructs skipped velocities by combining calibrated scalar statistics with each sample's orthogonal direction.

  \item \textbf{Strong speed-quality trade-offs across T2I and T2V.}
  In the main comparisons across BAGEL, FLUX.1-dev, and Wan2.1-1.3B, TACache achieves up to $4.11\times$ text-to-image speedup and $2.11\times$ text-to-video speedup. Compared with state-of-the-art caching methods, TACache achieves superior performance on most metrics, offering a better speed-quality trade-off.
\end{itemize}

\vspace{-2mm}
\section{Related Work}
\vspace{-2mm}
\subsection{Acceleration of diffusion and rectified flow inference}
The high inference cost of diffusion and RF models has motivated several complementary acceleration strategies. The first relies on distillation, e.g., consistency models~\cite{song2023consistency,lcm} and progressive distillation~\cite{salimans2022progressive}, which train a student model to reduce the number of sampling steps; this yields large speedups but requires additional training. The second improves the numerical solving process under a fixed pretrained model: DPM-Solver~\cite{dpm} and EDM~\cite{karras2022edm} use higher-order ODE solvers and design-space refinements to reduce coarse-step discretization error. A2S~\cite{a2s} augments the Euler update with a second-order acceleration correction, projects the spatial component onto the current velocity direction, and keeps only a magnitude-scaling term; this targets single-step Euler correction with a model evaluation at every step, whereas our method retains both the parallel component and the orthogonal residual for cache-based step skipping. A third line modifies training to straighten transport dynamics for coarse discretization, e.g., Reflow~\cite{reflow}, PeRFlow~\cite{perflow}, and ProReflow~\cite{proreflow}; in particular, ProReflow emphasizes the importance of direction matching over magnitude matching during training, whereas our decomposition is applied at inference time without retraining. Our work belongs to a fourth line, cache-based acceleration, which leverages the temporal redundancy of sampling by reusing earlier computations. As a training-free inference-side acceleration method, it preserves the underlying solver and is composable with the strategies above.
\vspace{-2mm}
\subsection{Cache-based acceleration}
\vspace{-2mm}
Cache-based methods address two central questions: \emph{when to skip} and \emph{how to approximate the skipped outputs}. TeaCache~\cite{teacache} decides when to skip online based on accumulated differences of timestep-embedding-modulated inputs rescaled by a fitted polynomial, and directly reuses cached outputs as the approximation; EasyCache~\cite{easycache} reuses cached transformation vectors and introduces a runtime-adaptive criterion based on cumulative-error monitoring to determine the caching schedule. TaylorSeer~\cite{taylorseer} adopts a relatively static caching schedule for the timing decision and approximates the skipped outputs via Taylor-series extrapolation of feature evolution. DiCache~\cite{dicache} uses online shallow-layer probes, exploiting the strong correlation between shallow-layer feature differences and deep-layer output differences, to dynamically determine the caching schedule and combines multi-step historical caches guided by the probe trajectory to approximate skipped steps. MagCache~\cite{magcache} fits an offline scalar magnitude curve, motivated by the monotonically decreasing ratio between consecutive residual norms, that serves both as the timing criterion and as a per-step scalar rescaling applied to historical outputs; observing that adjacent residuals differ primarily in magnitude, it focuses on the magnitude component and does not explicitly reconstruct the directional component. By contrast, our method retains the directional component and applies independent thresholds to the magnitude and direction indicators on both the \emph{when} and \emph{how} sides.

\begin{figure*}[!t]
  \centering
  \includegraphics[width=\textwidth]{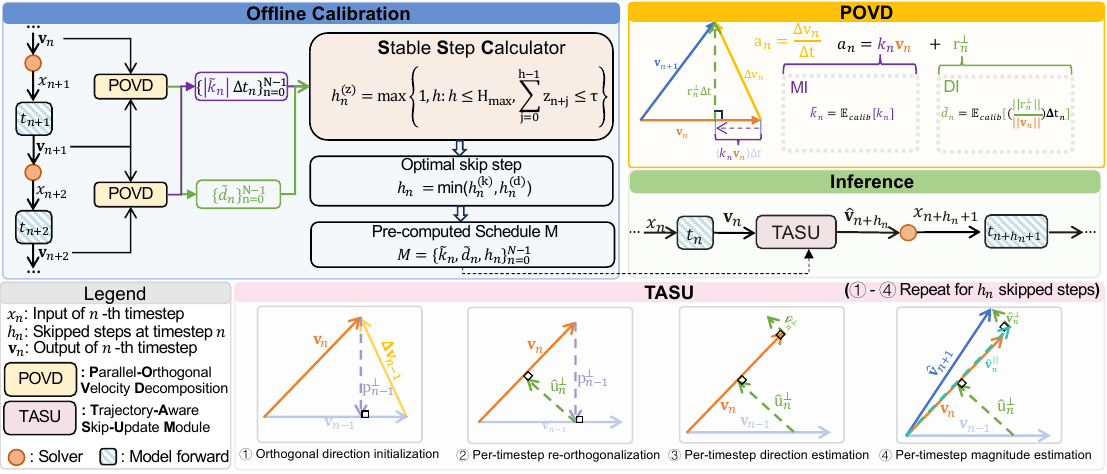}
  \vspace{-3mm}
  \caption{
    \textbf{Overview of TACache.} TACache consists of an offline calibration stage and an online inference stage. The calibration stage applies POVD to extract per-timestep statistics $(\tilde{k}_n, \tilde{d}_n)$, from which SSC constructs a fixed skip schedule; at inference, TASU reconstructs the velocity at skipped steps in four steps, fusing the offline statistics with each sample's own historical orthogonal direction. 
    }
  \label{fig:overview}
  \vspace{-4mm}
\end{figure*}

\vspace{-4mm}
\section{Method}
\vspace{-3mm}
\subsection{Preliminaries}
\label{sec:prelim}

\noindent\textbf{Flow Matching and Rectified Flow.}
Flow Matching~\cite{fm} learns a time-dependent vector field $v_\theta(\mathbf{X}, t, c)$ that approximates a target velocity field generating a prescribed probability path $\{p_t\}_{t \in [0,1]}$, with $c$ denoting conditioning information. Rectified Flow~\cite{rf} specializes this framework by straightening transport between noise and data along a linear path. We adopt the convention that $t = 1$ corresponds to the noise level and $t = 0$ to the data level, so that the linear path $\mathbf{X}_t = t \mathbf{X}_1 + (1 - t) \mathbf{X}_0$ has the constant target velocity $u_t = \mathbf{X}_1 - \mathbf{X}_0$, where $\mathbf{X}_1 \sim \mathcal{N}(0, \mathbf{I})$ and $\mathbf{X}_0$ is a clean data sample. The neural vector field $v_\theta$ is trained to approximate this target velocity.

\noindent\textbf{Euler sampler.}
Sampling integrates the ODE $\frac{d\mathbf{X}_t}{dt} = v_\theta(\mathbf{X}_t, t, c)$ from $t_0 = 1$ to $t_N = 0$ on a monotonically decreasing time grid $\{t_n\}_{n=0}^{N}$. A first-order Euler step yields
\begin{equation}
\label{eq:euler}
\mathbf{X}_{t_{n+1}} \approx \mathbf{X}_{t_n} - \Delta t_n \, v_\theta(\mathbf{X}_{t_n}, t_n, c), \qquad \Delta t_n \coloneqq t_n - t_{n+1} > 0,
\end{equation}
which advances the sample from noise $\mathbf{X}_{t_0}$ toward data $\mathbf{X}_{t_N}$ over $N$ steps. For brevity we write $\mathbf{v}_n \coloneqq v_\theta(\mathbf{X}_{t_n}, t_n, c)$ when the context is clear.
\vspace{-2mm}
\subsection{Overall Pipeline}
\label{sec:pipeline}

The TACache pipeline (Fig.~\ref{fig:overview}) consists of an offline calibration stage and an online inference stage linked by three modules. \textbf{POVD} decomposes the discrete acceleration along the current velocity into a parallel component and an orthogonal residual. The former captures magnitude evolution, while the latter captures directional change, yielding two sample-specific scalars $(k_n,d_n)$ that summarize per-timestep velocity dynamics. During offline calibration, these scalars are aggregated across a calibration set into the indicators $(\widetilde{k}_n, \widetilde{d}_n)$, which \textbf{SSC} consumes to produce a skip schedule $\mathcal{H} = \{h_n\}$. This schedule prescribes, at each step, the admissible skip interval length within which subsequent model evaluations may be skipped. At inference, the schedule drives \textbf{TASU}: at every skipped step it reconstructs the velocity by combining the cached indicators $(\widetilde{k}_n, \widetilde{d}_n)$ with each sample's own historical orthogonal direction along the same parallel-orthogonal axes introduced by POVD. The trajectory then returns to the standard solver at the next evaluated step.
\vspace{-2mm}
\subsection{Calibration Decomposition}
\label{sec:calibration}
\noindent\textbf{Parallel-Orthogonal Velocity Decomposition (POVD).}
POVD provides the geometric foundation of TACache, on which SSC and TASU build. We define the \emph{discrete acceleration} $\mathbf{a}_n$ as the finite difference of velocities between consecutive timesteps:
\begin{equation}
\label{eq:acc}
\mathbf{a}_n \coloneqq \frac{\mathbf{v}_{n+1} - \mathbf{v}_n}{\Delta t_n},
\end{equation}
which is obtained directly from sampler outputs and incurs no additional model evaluation. We decompose $\mathbf{a}_n$ along the current velocity $\mathbf{v}_n$ into components parallel and orthogonal to $\mathbf{v}_n$:
\begin{equation}
\label{eq:decomp}
\mathbf{a}_n = k_n \mathbf{v}_n + \mathbf{r}_n^{\perp}, \qquad k_n \coloneqq \frac{\langle \mathbf{a}_n, \mathbf{v}_n \rangle}{\|\mathbf{v}_n\|^2}, \qquad \mathbf{r}_n^{\perp} \coloneqq \mathbf{a}_n - k_n \mathbf{v}_n.
\end{equation}
The scalar $k_n$ quantifies the relative rate at which $\|\mathbf{v}_n\|$ changes along the trajectory and acts as a pure \emph{magnitude} signal; the residual $\mathbf{r}_n^{\perp}$, by construction orthogonal to $\mathbf{v}_n$, captures the intrinsic directional shift of the velocity field. To place the two signals on a common dimensionless scale, we define a directional change score $d_n$ that factors out both the local velocity magnitude and the integration step size, yielding a unitless measure of how much the velocity direction turns per step:
\begin{equation}
\label{eq:dir-score}
d_n \coloneqq \frac{\|\mathbf{r}_n^{\perp}\|}{\|\mathbf{v}_n\|} \Delta t_n.
\end{equation}

The per-sample scalars $k_n$ and $d_n$ exhibit dual stability across samples and along the sampling trajectory (Fig.~\ref{fig:MI-DI}), which justifies a faithful offline characterization that captures their behavior across the calibration distribution. We therefore aggregate $k_n$ and $d_n$ over a small calibration set $\mathcal{C}$ to obtain the \emph{magnitude indicator} (MI) $\widetilde{k}_n$ and \emph{direction indicator} (DI) $\widetilde{d}_n$, defined as the per-step expectations
\begin{equation}
\label{eq:mi-di}
\mathrm{MI}_n = \widetilde{k}_n \coloneqq \mathbb{E}_{\mathcal{C}}[k_n], \qquad \mathrm{DI}_n = \widetilde{d}_n \coloneqq \mathbb{E}_{\mathcal{C}}[d_n].
\end{equation}
This stability across samples justifies caching $(\widetilde{k}_n, \widetilde{d}_n)$ as population-level indicators, whereas the orthogonal residual $\mathbf{r}_n^{\perp}$ is sample-specific and reconstructed online at inference (Sec.~\ref{sec:tasu}).
\noindent\textbf{Stable Step Calculator (SSC).}
SSC turns the calibrated indicators $(\widetilde{k}_n, \widetilde{d}_n)$ into an offline skip schedule that determines, for each timestep $n$, the admissible skip interval length $h_n \geq 1$, where $h_n = 1$ denotes a standard one-step update and $h_n > 1$ specifies a skip interval over which the next $h_n - 1$ model evaluations are skipped.
Formally, given a non-negative per-step variation sequence
$\mathbf{z} = \{z_m\}_{m=0}^{N-1}$, a starting index $n$, and a tolerance
$\tau$, the function $\mathrm{SSC}(\mathbf{z}, n, \tau)$ returns the largest
admissible skip interval length whose cumulative variation stays within $\tau$:
\begin{equation}
\label{eq:ssc-single}
\mathrm{SSC}(\mathbf{z}, n, \tau)
\coloneqq
\max\Bigl\{
h \in \{1,2,\ldots,\min(H_{\max},\, N-n)\}
\;\Big|\;
h=1 \ \vee\ 
\sum_{j=0}^{h-1} z_{n+j} \le \tau
\Bigr\}.
\end{equation}
where $H_{\max}$ is the global maximum skip length, and the bound is tightened to $N-n$ near the end of the sequence to keep the cumulative sum within the available timesteps. The condition $h=1$ guarantees a standard one-step update when no skip interval satisfies the cumulative-variation constraint, ensuring SSC returns a well-defined interval length at every step.
The two POVD terms evolve at different rates and exhibit different degrees of stability (Fig.~\ref{fig:MI-DI}), so a single coupled threshold would be either too loose for the more stable term or too tight for the more variable one. To pre-compute a single schedule reused across all inference calls, SSC therefore applies Eq.~\eqref{eq:ssc-single} independently to the per-step magnitude sequence $\mathbf{K}\coloneqq\{|\widetilde{k}_m|\Delta t_m\}_{m=0}^{N-1}$ and the per-step direction sequence $\mathbf{D}\coloneqq\{\widetilde{d}_m\}_{m=0}^{N-1}$, with tolerances $\tau_k$ and $\tau_d$, yielding
\begin{equation}
\label{eq:ssc-pair}
h_n^{(k)} = \mathrm{SSC}(\mathbf{K}, n, \tau_k), \qquad
h_n^{(d)} = \mathrm{SSC}(\mathbf{D}, n, \tau_d).
\end{equation}
The final skip schedule $\mathcal{H}=\{h_n\}_{n=0}^{N-1}$ is obtained
by enforcing both constraints:
\begin{equation}
\label{eq:ssc-final}
h_n = \min\!\bigl(h_n^{(k)}, h_n^{(d)}\bigr).
\end{equation}
Thus, each interval is locally bounded by the more constraining component: $\tau_d$ limits directional drift, while $\tau_k$ limits magnitude variation. The per-step error decomposition in Sec~\ref{sec:bound} further clarifies this design by separating the parallel magnitude, orthogonal strength, and direction alignment terms. Bundling the calibrated indicators with the resulting admissible skip interval lengths for all subsequent inference-time skip decisions yields the pre-computed schedule $\mathcal{M} = \{\widetilde{k}_n, \widetilde{d}_n, h_n\}_{n=0}^{N-1}$, which is loaded once and reused across the inference stage.

\vspace{-2mm}
\subsection{Online Inference}
\label{sec:tasu}
\noindent\textbf{Update Rule under Cached Statistics.}
At inference, the online module \textbf{Trajectory-Aware Skip-Update (TASU)} reconstructs the velocity at skipped steps from cached statistics $(\widetilde{k}_n, \widetilde{d}_n)$, avoiding model evaluations within each skip interval. Substituting the POVD decomposition in Eq.~\eqref{eq:decomp} into the finite-difference velocity relation in Eq.~\eqref{eq:acc} yields
\begin{equation}
\label{eq:euler-decomp}
\mathbf{v}_{n+1} = (1 + k_n \Delta t_n)\, \mathbf{v}_n + \mathbf{r}_n^{\perp}\, \Delta t_n.
\end{equation}

The factor $(1 + k_n \Delta t_n)$ matches the first-order Taylor expansion of $\exp(k_n \Delta t_n)$. We therefore adopt the exponential form for the parallel component, which composes multiplicatively across multiple consecutive skipped steps and avoids the accumulation of Taylor truncation error:
\begin{equation}
\label{eq:update}
\mathbf{v}_{n+1} \approx \exp(k_n \Delta t_n) \mathbf{v}_n + d_n \|\mathbf{v}_n\| \mathbf{u}_n^{\perp},
\end{equation}
where $\mathbf{u}_n^{\perp} \coloneqq \mathbf{r}_n^{\perp} / \|\mathbf{r}_n^{\perp}\|$ is the unit orthogonal direction, and the orthogonal strength is rewritten through the directional change score $d_n$ from Eq.~\eqref{eq:dir-score}. Since $(k_n, d_n)$ depend on the next-step velocity $\mathbf{v}_{n+1}$, which is unavailable at skipped steps, we substitute the cached pair $(\widetilde{k}_n, \widetilde{d}_n)$, while the remaining quantity $\mathbf{u}_n^{\perp}$ is recovered from the sample's own history as described next.

We initialize the orthogonal direction from the most recent evaluated step. With both $\mathbf{v}_{n-1}$ and $\mathbf{v}_n$ available at the start of each skip interval, we compute the velocity increment $\Delta \mathbf{v}_{n-1} \coloneqq \mathbf{v}_n - \mathbf{v}_{n-1}$ and extract its component orthogonal to $\mathbf{v}_{n-1}$:
\begin{equation}
\label{eq:dir-init}
\mathbf{p}_{n-1}^{\perp} \coloneqq \Delta \mathbf{v}_{n-1} - \operatorname{proj}_{\mathbf{v}_{n-1}}(\Delta \mathbf{v}_{n-1}).
\end{equation}
The vector $\mathbf{p}_{n-1}^{\perp}$ encodes the local turning direction inferred from the last evaluated step and serves as the initial orthogonal anchor for the upcoming skip interval of length $h_n$ starting at index $n$.

Within a skip interval, $\mathbf{p}_{n-1}^{\perp}$ is constructed as 
orthogonal to $\mathbf{v}_{n-1}$ and is generally not orthogonal to the 
evolving reconstruction $\widehat{\mathbf{v}}_{n+j}$. To preserve the 
parallel-orthogonal decomposition along the trajectory, we orthogonalize 
$\mathbf{p}_{n-1}^{\perp}$ against $\widehat{\mathbf{v}}_{n+j}$ at each step,
\begin{equation}
\label{eq:dir-realign}
\widehat{\mathbf{u}}_{n+j}^{\perp} = \frac{\mathbf{p}_{n-1}^{\perp} 
- \operatorname{proj}_{\widehat{\mathbf{v}}_{n+j}}(\mathbf{p}_{n-1}^{\perp})}
{\big\| \mathbf{p}_{n-1}^{\perp} 
- \operatorname{proj}_{\widehat{\mathbf{v}}_{n+j}}(\mathbf{p}_{n-1}^{\perp}) 
\big\|},
\end{equation}
and substitute the cached statistics $(\widetilde{k}_{n+j}, \widetilde{d}_{n+j})$ 
into Eq.~\eqref{eq:update} to obtain the recursive update
\begin{equation}
\label{eq:tasu-rec}
\widehat{\mathbf{v}}_{n+j+1} \coloneqq 
\underbrace{\exp\!\left(\widetilde{k}_{n+j} \Delta t_{n+j}\right) 
\widehat{\mathbf{v}}_{n+j}}_{\text{magnitude update}} 
+ \underbrace{\widetilde{d}_{n+j} \|\widehat{\mathbf{v}}_{n+j}\| 
\widehat{\mathbf{u}}_{n+j}^{\perp}}_{\text{direction update}},
\end{equation}
seeded with $\widehat{\mathbf{v}}_n = \mathbf{v}_n$ at the start of the 
interval. We iterate Eq.~\eqref{eq:tasu-rec} for $j = 0, 1, \ldots, h_n - 1$, 
propagating the sample state through Eq.~\eqref{eq:euler} with the 
reconstructed velocities at each step until the next evaluated step, 
incurring no model evaluations. At step $n+h_n$, a standard model 
evaluation refreshes the velocity and starts the next scheduled interval, 
resetting the parallel-orthogonal axes and confining direction-alignment 
error within each interval.

\vspace{-2mm}
\subsection{Per-Step Error Decomposition and Control}
\label{sec:bound}
To analyze the approximation quality of TASU, we compare its update at a single skipped step with the \emph{oracle component-wise update}, the velocity that the same component-wise formula would produce if the sample-specific scalars and the true orthogonal direction were available:
\begin{equation}
\label{eq:oracle}
\mathbf{v}_{n+1}^{\star} \coloneqq \exp(k_n \Delta t_n) \mathbf{v}_n + d_n \|\mathbf{v}_n\| \mathbf{u}_n^{\perp}.
\end{equation}
Specializing the recursive TASU update of Eq.~\eqref{eq:tasu-rec} to a single step ($j=0$) gives
\begin{equation}
\label{eq:tasu-single}
\widehat{\mathbf{v}}_{n+1} \coloneqq \exp(\widetilde{k}_n \Delta t_n) \mathbf{v}_n + \widetilde{d}_n \|\mathbf{v}_n\| \widehat{\mathbf{u}}_n^{\perp},
\end{equation}
where $\widehat{\mathbf{u}}_n^{\perp}$ is the recursively re-orthogonalized historical estimate of $\mathbf{u}_n^{\perp}$ from Eq.~\eqref{eq:dir-realign}.

Applying the mean value theorem to the parallel term and an expansion of the orthogonal term yields
\begin{equation}
\label{eq:total-bound}
\frac{\big\| \widehat{\mathbf{v}}_{n+1} - \mathbf{v}_{n+1}^{\star} \big\|}{\|\mathbf{v}_n\|} \leq \sqrt{ C_n^2 |\widetilde{k}_n - k_n|^2 + (\widetilde{d}_n - d_n)^2 + 2 \widetilde{d}_n d_n (1 - \cos\theta_n) },
\end{equation}
with $C_n \coloneqq \Delta t_n \exp\!\big( \max(\widetilde{k}_n, k_n) \Delta t_n \big)$ and $\cos\theta_n \coloneqq \langle \widehat{\mathbf{u}}_n^{\perp}, \mathbf{u}_n^{\perp} \rangle$. The bound separates the local error into three interpretable factors: the magnitude error on the parallel axis, the orthogonal strength error on the orthogonal axis, and the direction alignment error from the historical direction estimate. SSC provides an actionable proxy for limiting the first two by enforcing the cumulative variation thresholds $\tau_k$ on $\mathbf{K}$ and $\tau_d$ on $\mathbf{D}$ within each skip interval, while the third is mitigated by the recursive re-orthogonalization in Eq.~\eqref{eq:dir-realign}. The full derivation is provided in Appendix~\ref{app:bound-proof}, and an empirical analysis of the trajectory-level behavior of this decomposition is deferred to Appendix~\ref{app:dir-error}.

\begin{figure*}[t]
  \centering
  \includegraphics[width=\textwidth]{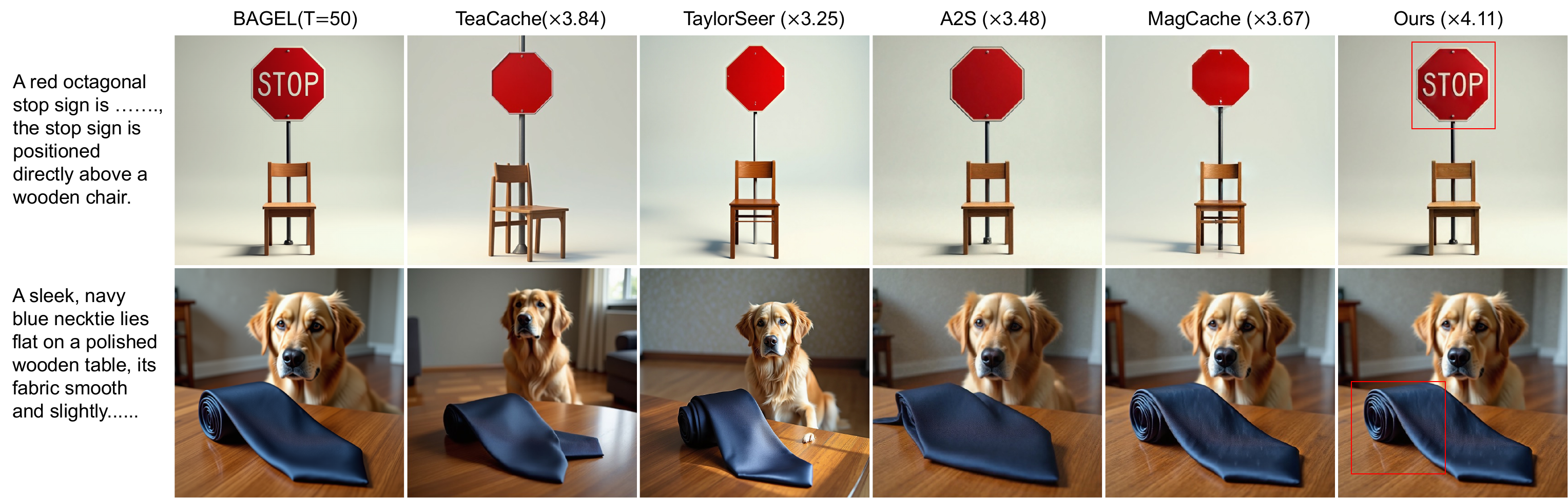}
  \vspace{-3mm}
  \caption{Visual quality comparison on BAGEL. Red boxes highlight where our method surpasses competing approaches with richer textural details and better semantic consistency.}
  \label{fig:bagel_visual}
  \vspace{-3mm}
\end{figure*}

\vspace{-3mm}
\section{Experiments}
\vspace{-3mm}
\label{sec:exp}
\subsection{Experimental Setting}
\vspace{-2mm}
\noindent\textbf{Models and Baselines.}
We evaluate TACache on three generative models: BAGEL~\cite{bagel} and FLUX.1-dev~\cite{flux} for T2I, and Wan2.1-1.3B~\cite{wan} for T2V. We compare TACache against representative acceleration baselines, including three cache-based methods, TeaCache~\cite{teacache}, TaylorSeer~\cite{taylorseer}, and MagCache~\cite{magcache}, as well as the acceleration-aware solver A2S~\cite{a2s}. We additionally report the full-step base models and a step-truncation baseline matched to our acceleration ratio.

\noindent\textbf{Benchmarks and Metrics.}
For T2I generation, we report results on GenEval~\cite{ghosh2023geneval}; for T2V generation, we report VBench~\cite{vbench}. To quantify visual preservation under acceleration, we treat the full-step base model outputs as references and report PSNR, SSIM~\cite{wang2004ssim}, and LPIPS~\cite{zhang2018lpips}, complemented by reference-free perceptual scores from CLIP-IQA~\cite{wang2023clipiqa} and Q-Align~\cite{wu2023qalign}. Efficiency is reported by per-sample latency, end-to-end speedup over the full-step base model, and PFLOPs per sample. More implementation details are provided in Appendix~\ref{app:implementation}.

\vspace{-2mm}
\subsection{Main Results}

\begin{figure*}[t!]
  \centering
  \includegraphics[width=\textwidth]{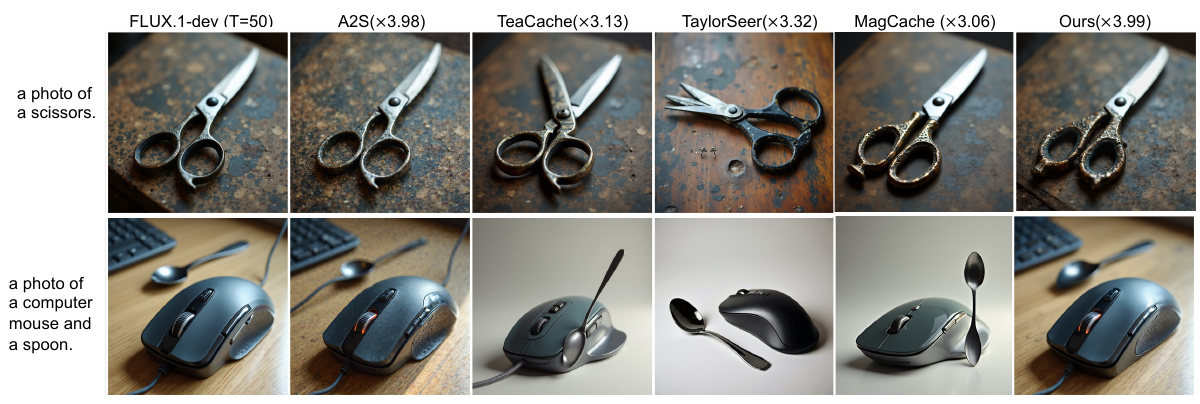}
  \vspace{-4mm}
  \caption{Qualitative comparison on FLUX.1-dev. The red boxes highlight defects in other methods, such as texture noise and severe structural distortions. Our method addresses these issues, generating high-fidelity images with coherent structures and clean textures. }
  \label{fig:flux_visual}
  \vspace{-4mm}
\end{figure*}

\begin{table*}[t!]
  \centering
    \caption{Quantitative evaluation of visual quality and inference efficiency on text-to-image generation. Our method attains the best quality--efficiency trade-off among all compared methods.}
  \label{tab:t2i}
  \setlength{\tabcolsep}{1.2mm}
  \renewcommand{\arraystretch}{1.05}
  \resizebox{\textwidth}{!}{%
  \begin{tabular}{lccccccccc}
    \toprule
    \multirow{2}{*}{\textbf{Method}}
      & \multicolumn{3}{c}{Efficiency}
      & \multicolumn{6}{c}{Visual Quality} \\
    \cmidrule(lr){2-4} \cmidrule(lr){5-10}
      & PFLOPs
      & Speedup
      & Latency (s)
      & GenEval
      & PSNR ($\uparrow$)
      & SSIM ($\uparrow$)
      & LPIPS ($\downarrow$)
      & CLIP-IQA ($\uparrow$)
      & Q-Align ($\uparrow$)  \\
    \midrule
      \multicolumn{10}{c}{\textbf{BAGEL (T=50)}} \\ % 加粗了子标题以示区分
    \midrule
    \rowcolor{gray!20}
    T=50
      & 5.25 & 1.00$\times$ & 51.13 & 85.91\%
      & $\infty$ & 1.0000 & 0.0000 & 0.8498 & 0.9489 \\
    T=13
      & \underline{1.29} & \underline{3.99$\times$} & \underline{12.83} & 85.01\%
      & 16.93 & 0.7943 & 0.2764 & 0.8052 & \underline{0.9404} \\
    TeaCache ~\cite{teacache}
      & \textbf{1.18} & 3.84$\times$ & 13.30 & 86.47\%
      & 20.80 & 0.8476 & 0.2047 & 0.7274 & 0.9291 \\
    TaylorSeer ~\cite{taylorseer}
      & \underline{1.29} & 3.25$\times$ & 15.73 & 82.12\%
      & 17.86 & 0.7860 & 0.2835 & \textbf{0.8351} & 0.9262 \\
    A2S ~\cite{a2s}$^*$
      & 1.60 & 3.48$\times$ & 14.67 & 87.27\%
      & 18.05 & 0.7386 & 0.4654 & 0.6183 & 0.8110 \\
    MagCache ~\cite{magcache}
      & 1.39 & 3.67$\times$ & 13.92 & \underline{87.29}\%
      & \underline{24.43} & \underline{0.8924} & \underline{0.1233} & 0.7690 & \textbf{0.9464} \\
    \rowcolor{orange!20}
    TACache (ours)
      & \underline{1.29} & \textbf{4.11$\times$} & \textbf{12.43} & \textbf{87.32}\%
      & \textbf{24.83} & \textbf{0.9049} & \textbf{0.1080} & \underline{0.8081} & 0.9315 \\
    \midrule
    \multicolumn{10}{c}{\textbf{FLUX.1-dev (T=50)}} \\ % 加粗了子标题以示区分
    \midrule
    \rowcolor{gray!20}
    T=50
      & 3.72 & 1.00$\times$ & 32.92 & 64.27\%
      &  $\infty$    & 1.0000    & 0.0000     & 0.9090 & 0.9623 \\
    T=13
      & \underline{0.98} & \textbf{3.99$\times$} & \textbf{8.26}  & \underline{63.87}\%
      & 16.11 & 0.6638 & 0.4321 & 0.8247 & 0.9495 \\
    TeaCache ~\cite{teacache}
      & 1.12 & 3.13$\times$ & 10.53 & 62.86\%
      & 15.49 & 0.6686 & 0.4282 & \underline{0.8952} & \underline{0.9603} \\
    TaylorSeer ~\cite{taylorseer}
      & 1.84 & 3.32$\times$ & 9.90  & 62.65\%
      & 9.20  & 0.3731 & 0.7530 & \textbf{0.9151} & \textbf{0.9694} \\
    A2S ~\cite{a2s}$^*$
      & \textbf{0.90} & \underline{3.98$\times$} & \underline{8.27}  & 59.01\%
      & \underline{18.41} & 0.6616 & 0.4752 & 0.7803 & 0.8662 \\
    MagCache ~\cite{magcache}
      & 1.25 & 3.06$\times$ & 10.76  & \textbf{64.61}\%
      & 16.49 & \underline{0.6970} & \underline{0.3845} & 0.8946 & 0.9582 \\
      \rowcolor{orange!20}
    TACache (ours)
      & \underline{0.98} & \textbf{3.99$\times$} & \textbf{8.26}  & 63.71\%
      & \textbf{22.57} & \textbf{0.8371} & \textbf{0.1969} & 0.8808 & 0.9370 \\
    \bottomrule
  \end{tabular}}
  \par\vspace{1mm}
  \parbox{\textwidth}{\footnotesize $^*$ A2S is reimplemented by us following the original paper.}
  \vspace{-2mm}
\end{table*}

\begin{table*}[t!]
\vspace{-2mm}
  \centering
  \caption{Quantitative evaluation on Wan2.1-1.3B. TACache achieves the fastest end-to-end latency among accelerated methods while maintaining competitive VBench performance.}
  \label{tab:wan}
  \setlength{\tabcolsep}{1.6mm}
  \renewcommand{\arraystretch}{1.05}
  \resizebox{\textwidth}{!}{%
  \begin{tabular}{lccccccccc}
    \toprule
    \multirow{2}{*}{Method} & \multicolumn{3}{c}{Efficiency} & \multicolumn{6}{c}{Visual Quality} \\
    \cmidrule(lr){2-4}\cmidrule(lr){5-10}
      & PFLOPs & Speedup & Latency (s) & VBench & PSNR ($\uparrow$) & SSIM ($\uparrow$) & LPIPS ($\downarrow$) & CLIP-IQA ($\uparrow$) & Q-Align ($\uparrow$) \\
    \midrule
     \rowcolor{gray!20}
    Wan2.1-1.3B            & 8.21 & 1.00$\times$ & 256.78 & 72.59\% & $\infty$   & 1.0000  & 0.0000  & 0.7118 & 0.9037 \\
    TeaCache~\cite{teacache}              & 4.11 & 1.91$\times$ & 134.57 & 72.38\% & 21.74      & 0.7686  & 0.1361  & \underline{0.6748} & 0.9016 \\
    TaylorSeer~\cite{taylorseer}           & \textbf{3.29} & \underline{2.10$\times$} & \underline{122.20} & 72.03\% & 14.48      & 0.4680  & 0.3877  & 0.6103 & 0.8800 \\
    MagCache~\cite{magcache}           & \underline{3.45} & 1.97$\times$ & 130.65 & \textbf{72.77}\% & \underline{23.82}      & \underline{0.8293}  & \underline{0.1065}  & 0.6687 & \underline{0.9024} \\
    \rowcolor{orange!20}
    TACache (ours)       & 3.78 & \textbf{2.11$\times$} & \textbf{121.60} & \underline{72.64}\% & \textbf{25.19}      & \textbf{0.8637}  & \textbf{0.0746}  & \textbf{0.7097} & \textbf{0.9096} \\
    \bottomrule
  \end{tabular}
  }
  \vspace{-2mm}
\end{table*}

\begin{figure*}[!t]
  \centering
  \includegraphics[width=\textwidth]{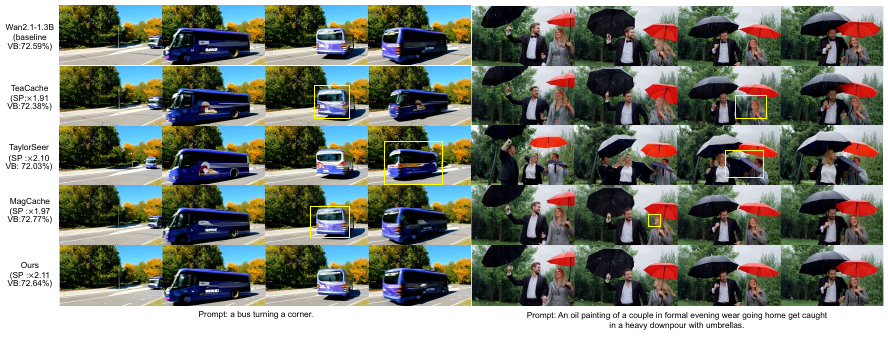}
  \vspace{-7mm}
  \caption{Qualitative results on Wan2.1-1.3B. Speedup and VBench score are reported for each method. Yellow boxes highlight texture collapse and motion incoherence in other methods.}
  \label{fig:wan_visual}
  \vspace{-6mm}
\end{figure*}

\vspace{-2mm}

\noindent\textbf{Quantitative Analysis.}
Tables~\ref{tab:t2i} and~\ref{tab:wan} compare TACache against accelerated baselines on BAGEL, FLUX.1-dev, and Wan2.1-1.3B~\cite{wan}. Across three backbones, TACache combines the highest acceleration with the strongest reference-based fidelity. On BAGEL it reaches a $4.11\times$ speedup with the best PSNR, SSIM, and LPIPS among accelerated methods, and a GenEval score of $87.32\%$ that slightly surpasses the full-step baseline of $85.91\%$. On FLUX.1-dev, it improves PSNR by more than $4$\,dB over the next-best accelerated method at a comparable $3.99\times$ speedup. On Wan2.1-1.3B, it attains the largest speedup of $2.11\times$ and the lowest latency of $121.60$\,s while raising PSNR from $23.82$\,dB under MagCache to $25.19$\,dB and keeping VBench close to MagCache and above the full-step baseline. Although non-reference metrics and benchmark scores such as CLIP-IQA, Q-Align, GenEval, and VBench occasionally peak at competing methods on individual settings, those methods consistently lag on reference-based fidelity, in line with our objective of preserving the base-model trajectory rather than optimizing non-reference scores alone. We attribute this balanced behavior to the parallel-orthogonal decomposition, which recovers both magnitude and direction at cached steps and preserves the full velocity trajectory; methods that restore only magnitude or directly reuse cached features tend to compromise either semantic alignment or pixel-level fidelity.

\begin{figure}[t]
\centering
\captionsetup[subfigure]{skip=4pt}

\begin{subfigure}{0.48\linewidth}
    \centering
    \includegraphics[width=\linewidth]{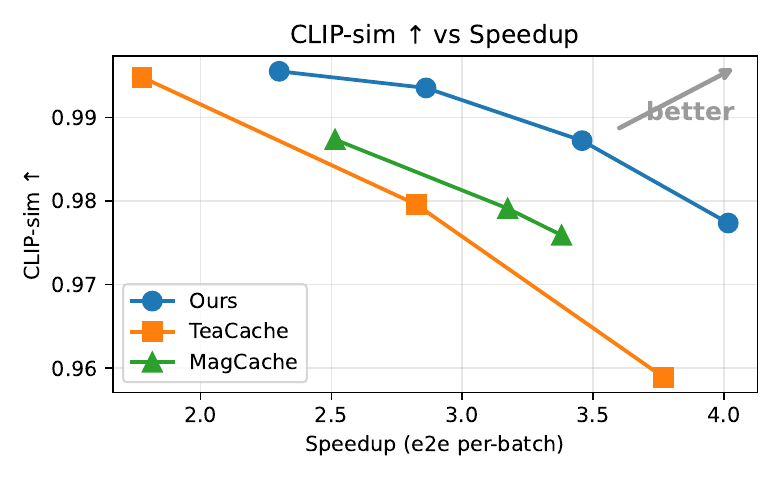}
    \caption{CLIP similarity vs.\ speedup.}
    \label{fig:pareto-clip}
\end{subfigure}
\hfill
\begin{subfigure}{0.48\linewidth}
    \centering
    \includegraphics[width=\linewidth]{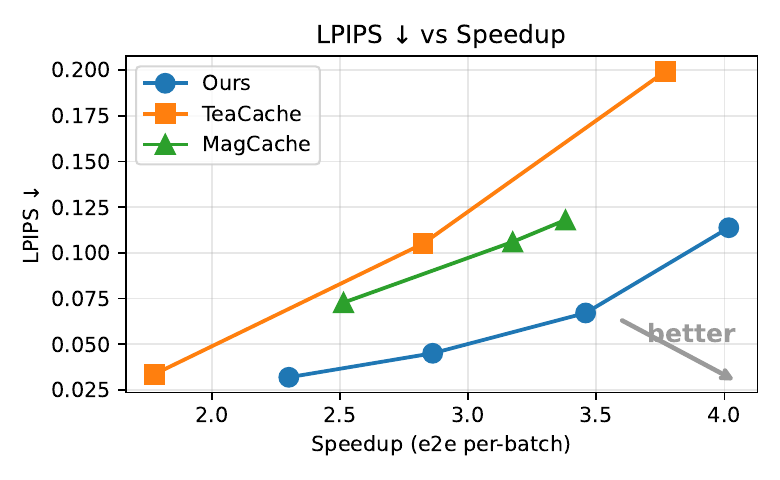}
    \caption{LPIPS vs.\ speedup.}
    \label{fig:pareto-lpips}
\end{subfigure}

\vspace{-2mm}
\caption{Pareto frontiers on BAGEL. TACache achieves a better speed-quality trade-off than TeaCache~\cite{teacache} and MagCache~\cite{magcache} on reference-free CLIP similarity and reference-based LPIPS.}
\label{fig:pareto}
\vspace{-2mm}
\end{figure}

\noindent\textbf{Pareto Frontier.}
TACache's advantage holds across operating points. As shown in Fig.~\ref{fig:pareto}, TACache traces a higher speed-quality frontier than TeaCache~\cite{teacache} and MagCache~\cite{magcache} on BAGEL. The gap becomes larger under aggressive acceleration: at about $3.4\times$ speedup, TACache reduces LPIPS to $0.067$, compared with $0.118$ for MagCache, while maintaining higher CLIP similarity. PSNR and SSIM frontiers in Appendix~\ref{app:pareto} show the same trend.

\vspace{-2mm}

\noindent\textbf{Qualitative Results.}
Figures~\ref{fig:bagel_visual}, \ref{fig:flux_visual}, and \ref{fig:wan_visual} compare TACache with accelerated baselines on the three backbones. A consistent pattern emerges: competing methods lose fine-grained detail under aggressive acceleration, whereas TACache remains visually close to the baseline reference. The failure modes differ across backbones, including missing textures and semantic cues on BAGEL, structural distortion of thin objects on FLUX.1-dev, and texture collapse with motion incoherence on Wan2.1-1.3B. TACache avoids these artifacts and better preserves spatial detail and temporal consistency. Additional comparisons are provided in Appendix~\ref{appfig}.

\vspace{-2mm}
\subsection{Ablation Study}
\vspace{-2mm}
\begin{table*}[t]
\centering
\begin{minipage}[t]{0.48\textwidth}
\centering
\caption{Component ablation on BAGEL. SSC improves over the step-truncation baseline; DI yields quality gains by constraining directional drift; TACache achieves the best GenEval.}
\label{tab:ablation1}
\scriptsize
\setlength{\tabcolsep}{1.3mm}
\renewcommand{\arraystretch}{1.2}
\begin{tabular}{ccccccc}
\toprule
SSC & MI & DI & GenEval & PSNR$\uparrow$ & Q-Align$\uparrow$ & CLIP-IQA$\uparrow$ \\
\midrule
\rowcolor{gray!20}
            &            &            & 85.01\% & 16.93 & 0.9404 & 0.8052 \\
\checkmark  &            &            & 86.16\% & 25.65 & 0.9229 & 0.7936 \\
\checkmark  & \checkmark &            & 86.06\% & 25.61 & 0.9210 & 0.7907 \\
\checkmark  &            & \checkmark & 87.21\% & 24.78 & 0.9341 & 0.8105 \\
\rowcolor{orange!20}
\checkmark  & \checkmark & \checkmark & 87.32\% & 24.83 & 0.9315 & 0.8081 \\
\bottomrule
\end{tabular}
\end{minipage}%
\hfill%
\begin{minipage}[t]{0.48\textwidth}
\centering
\caption{Speed-quality trade-off under varying thresholds. Moderate $\tau_k$, $\tau_d$ form a stable region; overly relaxed directional constraints degrade fidelity. Latency reported in seconds.}
\label{tab:tau-ablation}
\scriptsize
\setlength{\tabcolsep}{1mm}
\renewcommand{\arraystretch}{1.2}
\begin{tabular}{ccccccc}
\toprule
$\tau_k$ & $\tau_d$ & Latency & GenEval & PSNR$\uparrow$ & LPIPS$\downarrow$ & CLIP-IQA$\uparrow$ \\
\midrule
\rowcolor{gray!20}
0.00 & 0.0 & 51.13 & 85.91\% & $\infty$ & 0.0000 & 0.8498 \\
0.05 & 0.6 & 13.16 & 87.05\% & 24.81    & 0.1108 & 0.8041 \\
0.05 & 0.9 & 10.62 & 85.17\% & 19.99    & 0.2758 & 0.7017 \\
\rowcolor{orange!20}
0.06 & 0.6 & 12.43 & 87.32\% & 24.83    & 0.1080 & 0.8081 \\
0.07 & 0.7 & 12.54 & 86.59\% & 23.20    & 0.1576 & 0.7896 \\
\bottomrule
\end{tabular}
\end{minipage}
\vspace{-4mm}
\end{table*}

We ablate TACache on BAGEL along two axes: per-component contribution and sensitivity to magnitude and direction thresholds. The robustness of offline calibration is examined in Appendix~\ref{calib}.

\vspace{-2mm}
\noindent\textbf{Module-wise Ablation.}
Table~\ref{tab:ablation1} isolates each component under comparable speedup. SSC raises GenEval from $85.01\%$ to $86.16\%$ and PSNR by over $8$~dB above the step-truncation baseline. MI alone offers no gain, indicating magnitude is not the main degradation source. DI provides the main correction, lifting GenEval to $87.21\%$ and improving CLIP-IQA. The complete TACache configuration reaches $87.32\%$ GenEval while preserving reconstruction fidelity, confirming the complementary roles of magnitude and direction. Full results in Appendix~\ref{app:full-ablation}.

\noindent\textbf{Threshold Sensitivity.}
Table~\ref{tab:tau-ablation} characterizes the speed-quality landscape over the magnitude threshold $\tau_k$ and the direction threshold $\tau_d$. Moderate thresholds form a stable operating region in which TACache reaches around $4\times$ speedup while surpassing the baseline, with the best setting reaching $87.32\%$ GenEval. Relaxing the direction threshold pushes the speedup further but drops GenEval below the baseline, indicating that direction control is essential for limiting trajectory drift under aggressive skipping. A finer sweep with full metrics is reported in Appendix~\ref{app:full-ablation}.

\vspace{-3mm}
\section{Conclusion}
\vspace{-3mm}
We presented TACache, which reframes timestep skipping in Rectified Flow models as a geometric trajectory reconstruction problem, recovering skipped dynamics through the magnitude and direction structure of velocity evolution. By calibrating stable magnitude and direction indicators offline and combining them with the latest velocity history online, TACache reconstructs skipped updates without additional model evaluations. Main comparisons on BAGEL, FLUX.1-dev, and Wan2.1-1.3B show speedups of up to $4.11\times$ on text-to-image generation and $2.11\times$ on text-to-video generation, with consistent improvements over prior cache-based methods on all reference-based fidelity metrics. These results suggest that preserving trajectory geometry offers an effective principle for cache-based diffusion acceleration. Future work will extend TACache to longer video generation and integrate it with distillation-based samplers to compound the acceleration gains.

\begin{ack}
This work was supported by Huawei Explore X funds. This work was also sponsored by AI Hundred Schools Program and was carried out using the Ascend AI technology stack.
\paragraph{Competing interests.}
The authors declare no additional competing interests beyond the affiliations and funding disclosed above.
\end{ack}

\bibliographystyle{plainnat}
\bibliography{ref} 

\newpage
\appendix
% \section*{Appendix}
\section{Implementation Details}
\label{app:implementation}

T2I images are generated at $1024 \times 1024$ resolution, while T2V clips are generated at $832 \times 480$ resolution with $81$ frames; each backbone uses its default classifier-free guidance scale. All accelerated baselines are run under their officially released configurations, and the step-truncation baseline matches our acceleration ratio with $T=13$ for T2I.

GenEval~\cite{ghosh2023geneval} evaluates fine-grained semantic alignment across multiple sub-tasks, with per-sub-task scores reported in Appendix~\ref{app:geneval}. Following previous work~\cite{vbench1,vbench2}, VBench~\cite{vbench} scores are aggregated over its eight dimensions. PFLOPs per sample include both model evaluations and solver overhead.

We construct calibration sets by sampling 60 prompts from WISE~\cite{wise} for the T2I models and 30 prompts from InternVid-10M-FLT~\cite{wang2024internvid} for the T2V model, filtered by aesthetic quality and prompt category; the sensitivity to calibration set size is reported in Table~\ref{tab:calibration}. At each sampling step, trajectory statistics are aggregated across calibration samples using the mean plus one standard deviation to construct the offline curves.
The maximum skip interval length is set to $H_{\max}=12$ for T2I models and $H_{\max}=3$ for the T2V model. The SSC thresholds $\tau_k$ and $\tau_d$ are set to $0.06$ and $0.6$ for BAGEL, $0.05$ and $0.5$ for FLUX.1-dev, and $0.03$ and $0.3$ for Wan2.1-1.3B; their sensitivity on BAGEL is analyzed in Table~\ref{tab:app_tau-ablation}. For Wan2.1-1.3B, we divide the latent temporal dimension into five contiguous blocks, using block-specific trajectory statistics for velocity compensation while determining a shared skip schedule from their averaged statistics.
The latency reported in Table~\ref{tab:t2i} and Table~\ref{tab:wan} is averaged over the GenEval and VBench prompt sets.
\vspace{-4mm}
\section{Pseudocode of TACache}
\vspace{-2mm}
\label{app:algorithm}

Algorithm~\ref{alg:tacache} summarizes TACache inference with the pre-computed schedule.

\begin{algorithm}[H]
\caption{TACache Inference with Offline Skip Schedule}
\label{alg:tacache}
\KwIn{Timesteps $\{t_n\}_{n=0}^{N}$, pre-computed schedule $\mathcal{M}=\{(\widetilde{k}_n,\widetilde{d}_n,h_n)\}_{n=0}^{N-1}$, model $v_\theta$, condition $c$}
\KwOut{Sample $\mathbf{X}_{t_N}$}

$\mathbf{X}_{t_0} \sim \mathcal{N}(0,\mathbf{I})$ \tcp*{noise initialization}
$n \gets 0$\;

\While{$n < N$}{
    $h \gets
    \begin{cases}
        1, & n < 2 \text{ or } n \geq N-2,\\
        \min(h_n, N-2-n), & \text{otherwise}
    \end{cases}$\;

    \eIf{$h = 1$}{
        $\mathbf{v}_n \gets v_\theta(\mathbf{X}_{t_n}, t_n, c)$\;
        $\mathbf{X}_{t_{n+1}} \gets \mathrm{Update}(\mathbf{X}_{t_n}, \mathbf{v}_n, t_n, t_{n+1})$\;
        $n \gets n + 1$\;
    }{
        $\mathbf{v}_n \gets v_\theta(\mathbf{X}_{t_n}, t_n, c)$ \tcp*{anchor evaluation}
        $\mathbf{p}_{n-1}^{\perp} \gets \mathrm{InitDirection}(\mathbf{v}_n, \mathbf{v}_{n-1})$\;
        $\widehat{\mathbf{v}}_n \gets \mathbf{v}_n$ \tcp*{seed TASU recursion}

        \For{$j \gets 0$ \KwTo $h-1$}{
            \If{$j < h-1$}{
                $\widehat{\mathbf{u}}_{n+j}^{\perp} \gets \mathrm{ReOrth}(\mathbf{p}_{n-1}^{\perp}, \widehat{\mathbf{v}}_{n+j})$\;
                $\widehat{\mathbf{v}}_{n+j+1} \gets \exp(\widetilde{k}_{n+j}\Delta t_{n+j}) \widehat{\mathbf{v}}_{n+j} + \widetilde{d}_{n+j} \|\widehat{\mathbf{v}}_{n+j}\| \widehat{\mathbf{u}}_{n+j}^{\perp}$\;
            }
            $\mathbf{X}_{t_{n+j+1}} \gets \mathrm{Update}(\mathbf{X}_{t_{n+j}}, \widehat{\mathbf{v}}_{n+j}, t_{n+j}, t_{n+j+1})$\;
            $\mathbf{v}_{n+j} \gets \widehat{\mathbf{v}}_{n+j}$ \tcp*{cache used velocity}
        }
        $n \gets n + h$\;
    }
}
\Return $\mathbf{X}_{t_N}$
\end{algorithm}

\noindent
The historical velocity $\mathbf{v}_{n-1}$ denotes the velocity used at the preceding sampling step, which may be either model-evaluated or predicted by TASU. This velocity is cached after each step and used for subsequent direction initialization. The first two and last two sampling steps are always evaluated normally. The function $\mathrm{InitDirection}$ implements Eq.~\eqref{eq:dir-init}, $\mathrm{ReOrth}$ implements Eq.~\eqref{eq:dir-realign}, and $\mathrm{Update}$ performs the corresponding solver step. Since SSC returns $h_n \geq 1$, $h_n=1$ denotes a standard one-step update and $h_n>1$ denotes a skip interval.

\section{Per-Step Error Bound and Trajectory Diagnostics}
\label{app:bound}

\subsection{Detailed Derivation of the Per-Step Error Bound}
\label{app:bound-proof}

We provide the full derivation of the per-step bound in Eq.~\eqref{eq:total-bound}. The main text outlines the key steps and intuition; here we present the complete mathematical details.

\noindent\textbf{Setup and assumptions.}
Throughout this derivation we assume $\|\mathbf{v}_n\| > 0$, $\|\mathbf{r}_n^\perp\| > 0$, and $\|\mathbf{p}_{n-1}^{\perp} - \operatorname{proj}_{\mathbf{v}_n}(\mathbf{p}_{n-1}^{\perp})\| > 0$, so that the true direction $\mathbf{u}_n^{\perp}$ defined in Sec~\ref{sec:tasu} and the historical estimate $\widehat{\mathbf{u}}_n^{\perp}$ obtained from Eq.~\eqref{eq:dir-realign} at $j = 0$ are both well-defined. At $j=0$, TASU is seeded with $\widehat{\mathbf{v}}_n=\mathbf{v}_n$, so the historical anchor is re-orthogonalized against $\mathbf{v}_n$. Degenerate cases with vanishing orthogonal components are rare in practice; when they occur, the corresponding directional update is set to zero. Recall also that $d_n \geq 0$ by Eq.~\eqref{eq:dir-score} and $\widetilde{d}_n \geq 0$ as the average of nonnegative quantities.

\noindent\textbf{Error decomposition.}
Recall the oracle update from Eq.~\eqref{eq:oracle} and the single-step TASU reconstruction from Eq.~\eqref{eq:tasu-single}:
\begin{align}
\mathbf{v}_{n+1}^\star &= \exp(k_n \Delta t_n) \mathbf{v}_n + d_n \|\mathbf{v}_n\| \mathbf{u}_n^{\perp}, \\
\widehat{\mathbf{v}}_{n+1} &= \exp(\widetilde{k}_n \Delta t_n) \mathbf{v}_n + \widetilde{d}_n \|\mathbf{v}_n\| \widehat{\mathbf{u}}_n^{\perp}.
\end{align}
Both updates are evaluated from the same current velocity $\mathbf{v}_n$, so the bound below isolates the local substitution error rather than accumulated trajectory drift. Their difference splits into a parallel component along $\mathbf{v}_n$ and a residual involving the orthogonal directions:
\begin{equation}
\widehat{\mathbf{v}}_{n+1} - \mathbf{v}_{n+1}^\star = \underbrace{\big[ \exp(\widetilde{k}_n \Delta t_n) - \exp(k_n \Delta t_n) \big] \mathbf{v}_n}_{P} + \underbrace{\|\mathbf{v}_n\| \big[ \widetilde{d}_n \widehat{\mathbf{u}}_n^{\perp} - d_n \mathbf{u}_n^{\perp} \big]}_{Q}.
\end{equation}

\noindent\textbf{Orthogonality of $P$ and $Q$.}
Both $\widehat{\mathbf{u}}_n^{\perp}$ and $\mathbf{u}_n^{\perp}$ lie in the subspace orthogonal to $\mathbf{v}_n$, the former by the construction of Eq.~\eqref{eq:dir-realign} and the latter by definition. Hence their linear combination $Q$ also lies in that subspace. Since $P$ is parallel to $\mathbf{v}_n$, we have $P \perp Q$, and the squared error decomposes additively:
\begin{equation}
\label{eq:appx-decomp}
\big\| \widehat{\mathbf{v}}_{n+1} - \mathbf{v}_{n+1}^\star \big\|^2 = \|P\|^2 + \|Q\|^2.
\end{equation}
Note that this decomposition does not require $\widehat{\mathbf{u}}_n^{\perp}$ and $\mathbf{u}_n^{\perp}$ to be orthogonal to each other; in general, they differ by an angle $\theta_n$ as defined in Sec~\ref{sec:bound}.

\noindent\textbf{Bounding $\|P\|^2$ via the mean value theorem.}
By the mean value theorem applied to $f(x) = \exp(x \Delta t_n)$, there exists $\xi$ between $k_n$ and $\widetilde{k}_n$ such that
\begin{equation}
\exp(\widetilde{k}_n \Delta t_n) - \exp(k_n \Delta t_n) = \Delta t_n \exp(\xi \Delta t_n) (\widetilde{k}_n - k_n).
\end{equation}
Recall the step-dependent constant $C_n \coloneqq \Delta t_n \exp\!\big( \max(\widetilde{k}_n, k_n) \Delta t_n \big)$ introduced in Sec~\ref{sec:bound}. Since $\Delta t_n > 0$ and $\xi \leq \max(\widetilde{k}_n, k_n)$, the monotonicity of $\exp(\cdot)$ gives $\exp(\xi \Delta t_n) \leq \exp(\max(\widetilde{k}_n, k_n) \Delta t_n)$, hence $\big| \exp(\widetilde{k}_n \Delta t_n) - \exp(k_n \Delta t_n) \big| \leq C_n |\widetilde{k}_n - k_n|$. Squaring and multiplying by $\|\mathbf{v}_n\|^2$ then gives
\begin{equation}
\label{eq:appx-P}
\|P\|^2 \leq C_n^2 |\widetilde{k}_n - k_n|^2 \|\mathbf{v}_n\|^2.
\end{equation}

\noindent\textbf{Expanding $\|Q\|^2$ exactly.}
Since $\widehat{\mathbf{u}}_n^{\perp}$ and $\mathbf{u}_n^{\perp}$ are unit vectors with $\langle \widehat{\mathbf{u}}_n^{\perp}, \mathbf{u}_n^{\perp} \rangle = \cos\theta_n$, we have
\begin{align}
\|Q\|^2
&= \|\mathbf{v}_n\|^2 \big\| \widetilde{d}_n \widehat{\mathbf{u}}_n^{\perp} - d_n \mathbf{u}_n^{\perp} \big\|^2 \notag \\
&= \|\mathbf{v}_n\|^2 \big( \widetilde{d}_n^2 + d_n^2 - 2 \widetilde{d}_n d_n \cos\theta_n \big) \notag \\
&= \|\mathbf{v}_n\|^2 \big[ (\widetilde{d}_n - d_n)^2 + 2 \widetilde{d}_n d_n (1 - \cos\theta_n) \big], \label{eq:appx-Q}
\end{align}
where the last step uses the algebraic identity $a^2 + b^2 - 2ab\cos\theta = (a - b)^2 + 2ab(1 - \cos\theta)$. This is an identity rather than a bound: the orthogonal-component error has no slack beyond the substitution itself.

\noindent\textbf{Combining and taking the square root.}
Substituting Eqs.~\eqref{eq:appx-P} and~\eqref{eq:appx-Q} into Eq.~\eqref{eq:appx-decomp} gives
\begin{equation}
\big\| \widehat{\mathbf{v}}_{n+1} - \mathbf{v}_{n+1}^\star \big\|^2 \leq \|\mathbf{v}_n\|^2 \big[ C_n^2 |\widetilde{k}_n - k_n|^2 + (\widetilde{d}_n - d_n)^2 + 2 \widetilde{d}_n d_n (1 - \cos\theta_n) \big].
\end{equation}
Dividing by $\|\mathbf{v}_n\|^2$ and taking the square root yields Eq.~\eqref{eq:total-bound}. Setting $\theta_n = 0$ eliminates the alignment term and recovers the scalar-substitution bound
\begin{equation}
\label{eq:scalar-bound}
\frac{\big\| \widehat{\mathbf{v}}_{n+1} - \mathbf{v}_{n+1}^\star \big\|}{\|\mathbf{v}_n\|} \leq \sqrt{ C_n^2 |\widetilde{k}_n - k_n|^2 + (\widetilde{d}_n - d_n)^2 },
\end{equation}
which depends only on the scalar-substitution gaps between the cached indicators $(\widetilde{k}_n, \widetilde{d}_n)$ and the sample-specific scalars $(k_n, d_n)$.

\subsection{Empirical Analysis of Trajectory-Level Behavior}
\label{app:dir-error}

To complement the per-step analysis in Sec~\ref{sec:bound}, we report the trajectory behavior of TACache and MagCache under their respective calibrated skip schedules. We run both methods on $50$ GenEval prompts under BAGEL, and at each timestep $n$ measure the relative state drift $\|\mathbf{X}_{t_n}^{\mathrm{cache}} - \mathbf{X}_{t_n}^{\mathrm{full}}\| / \|\mathbf{X}_{t_n}^{\mathrm{full}}\|$, the relative velocity drift $\|\mathbf{v}_n^{\mathrm{cache}} - \mathbf{v}_n^{\mathrm{full}}\| / \|\mathbf{v}_n^{\mathrm{full}}\|$, and the cosine alignment $\cos\theta_n$ between the historical direction estimate and the oracle orthogonal direction defined in Sec~\ref{sec:tasu} and used in Sec~\ref{sec:bound}. The diagnostic compares per-step quantities along two trajectories advanced from the same noise initialization, and is therefore distinct from the same-state local bound derived in Appendix~\ref{app:bound-proof}.

\begin{figure}[t]
\centering
\includegraphics[width=0.85\linewidth]{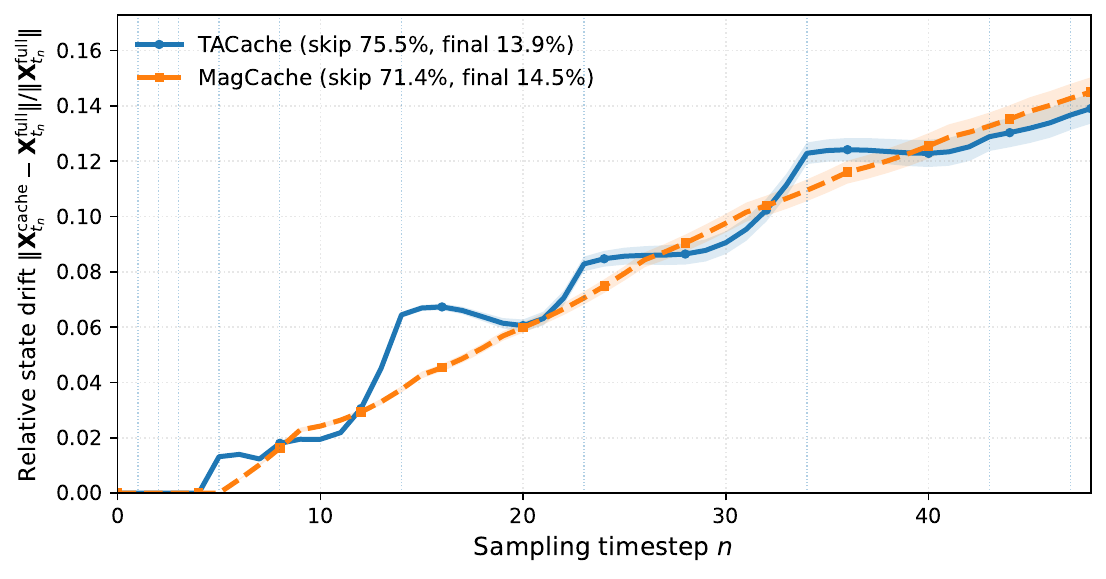}
\caption{Mean relative state drift along the $50$-step sampling trajectory on $50$ GenEval prompts under BAGEL. Both methods are evaluated against the corresponding full-step trajectory advanced from the same noise initialization. TACache reaches a final state drift of $13.9\%$ at a $75.5\%$ skip ratio, while MagCache reaches $14.5\%$ at a $71.4\%$ skip ratio. Shaded bands denote the standard error across prompts; vertical dotted lines mark TACache's scheduled evaluated steps.}
\label{fig:trajectory_error}
\end{figure}

\noindent\textbf{Trajectory drift profile.}
Figure~\ref{fig:trajectory_error} reports the mean relative state drift across $50$ prompts as a function of timestep. Under its calibrated skip schedule, TACache skips $75.5\%$ of the sampling steps and reaches a final state drift of $13.9\%$, while MagCache skips $71.4\%$ and reaches $14.5\%$. Scheduled evaluated steps reduce the velocity drift relative to cached steps, thereby interrupting the accumulation induced by consecutive cached updates.

\noindent\textbf{Cached versus evaluated steps.}
Table~\ref{tab:skip_forward} reports the per-step velocity drift on the two subsets. Velocity drift on cached steps is markedly higher than on evaluated steps for both methods: TACache averages $36.2\%$ on cached steps and $17.0\%$ on evaluated steps, while MagCache averages $20.2\%$ and $10.9\%$, respectively. TACache exhibits a higher per-step drift on cached steps because of its more aggressive skip schedule, but its final state drift remains lower thanks to the timely anchoring at scheduled evaluated steps. The contrast confirms that evaluated steps embedded in the schedule act as drift anchors, in line with the role of $\tau_k$ and $\tau_d$ in Eq.~\eqref{eq:ssc-pair}, which limit how far each skip interval may extend before a new model evaluation is enforced.

\noindent\textbf{Direction alignment.}
We further examine the historical direction estimate that drives the alignment term $2 \widetilde{d}_n d_n (1 - \cos\theta_n)$ in Eq.~\eqref{eq:total-bound}. On cached steps of TACache, $\cos\theta_n$ has mean $0.044$ and is positive in $89.3\%$ of records, with the $90$-th percentile at $0.091$. The estimate is far from a perfect match to the oracle direction, but it provides a weak yet consistently positive alignment signal after recursive re-orthogonalization. Together with the calibrated direction indicator $\widetilde{d}_n$, this alignment signal helps limit the exposure to direction mismatch in cached updates and complements the cumulative-variation control on the magnitude and orthogonal-strength terms.

\begin{table}[t]
\centering
\caption{Per-step velocity drift on cached and evaluated-step subsets across $50$ GenEval prompts on BAGEL. Each entry reports the mean relative $\ell_2$ error $\|\mathbf{v}^{\mathrm{cache}} - \mathbf{v}^{\mathrm{full}}\| / \|\mathbf{v}^{\mathrm{full}}\|$.}
\label{tab:skip_forward}
 \vspace{2mm}
\setlength{\tabcolsep}{6pt}
\renewcommand{\arraystretch}{1.15}
\small
\begin{tabular}{l c c c c}
\hline
\textbf{Method} & \textbf{Skip ratio} & \textbf{Cached steps} & \textbf{Evaluated steps} & \textbf{Final state drift} \\
\hline
TACache  & $75.5\%$ & $0.362$ & $0.170$ & $0.139$ \\
MagCache & $71.4\%$ & $0.202$ & $0.109$ & $0.145$ \\
\hline
\end{tabular}
\end{table}

\vspace{-2mm}
\section{Calibration Efficiency}
\label{calib}
\vspace{-2mm}
\begin{table}[t!]
  \centering
  \caption{Calibration efficiency on BAGEL. TACache remains effective with small calibration sets, while larger sets modestly improve quality at higher offline cost.}
  \label{tab:calibration}
  \setlength{\tabcolsep}{4.4mm}
  \renewcommand{\arraystretch}{1.05}
  \resizebox{\linewidth}{!}{%
  \begin{tabular}{lccccc}
    \toprule
    Source & Number & Offline Time (min) & GenEval & Latency (s) & Speedup \\
    \midrule
    \rowcolor{gray!20}
    Baseline (full-step) & --- & --- & 85.91\% & 51.13 & 1.00$\times$ \\
    WISE & 6 & 9.3 & 86.60\% & 11.48 & 4.45$\times$ \\
    WISE & 30 & 31.0 & 87.14\% & 11.48 & 4.45$\times$ \\
    \rowcolor{orange!20}
    WISE & 60 & 58.3 & \textbf{87.32\%} & 12.43 & 4.11$\times$ \\
    CompBench & 60 & 58 & 86.60\% & 12.35 & 4.14$\times$ \\
    \bottomrule
  \end{tabular}%
  }
  \vspace{-2mm}
\end{table}

Table~\ref{tab:calibration} examines the offline calibration cost and its effect on final generation quality. With WISE prompts, increasing the calibration size from $6$ to $60$ improves GenEval from $86.60\%$ to $87.32\%$, with offline cost rising from under $10$ minutes to about one hour, while the inference speedup remains stable between $4.11\times$ and $4.45\times$. Notably, using only $6$ prompts already yields a GenEval score above the baseline of $85.91\%$, indicating that the trajectory statistics can be reliably estimated even from a small calibration set. Replacing WISE with CompBench under the same $60$-prompt budget gives a comparable GenEval score of $86.60\%$, suggesting that the calibration is largely robust to the choice of prompt source. Overall, these results show that TACache requires only lightweight offline calibration, without requiring any additional inference-time computation, and continues to benefit from larger calibration sets when additional offline cost is acceptable in deployment.

\vspace{-2mm}
\section{Full Ablation Results}
\label{app:full-ablation}
\vspace{-2mm}
We report the full per-metric breakdown for the two ablation studies summarized in the main paper. Table~\ref{tab:app_ablation1} extends the component ablation with three additional metrics, namely Accuracy, SSIM, and LPIPS, where Accuracy denotes the per-image correctness rate produced by the GenEval pipeline and provides a finer view of prompt-image alignment than the aggregated GenEval score. Table~\ref{tab:app_tau-ablation} provides the corresponding full threshold sweep, including speedup, SSIM, and Q-Align. The conclusions reported in the main paper are preserved across all $7$ metrics. SSC alone, by replacing the step-truncation baseline with the calibrated skip schedule, improves both reference-based fidelity and GenEval. MI is designed as a scale stabilizer that complements DI rather than a standalone module, since the global magnitude trend has already been absorbed by SSC during offline calibration; its role is to keep the magnitude of cached updates aligned with the full-step reference once DI introduces directional corrections. DI delivers the dominant correction in the inference loop, lifting GenEval, Accuracy, and CLIP-IQA above the SSC baseline. The complete TACache configuration achieves the best Accuracy at $86.93\%$ alongside the best GenEval, and remains close to the strongest accelerated variant under the same speedup setting on SSIM and Q-Align. Although CLIP-IQA peaks at the $\mathrm{SSC}{+}\mathrm{DI}$ configuration, the complete configuration stays within $0.003$ of this peak, suggesting that incorporating MI introduces only a small trade-off.

\begin{table*}[t]
  \centering
  \caption{Component ablation on BAGEL. The first row denotes the step-truncation baseline. SSC improves over this baseline, DI provides the main quality gain by constraining directional drift, and the complete TACache configuration achieves the best GenEval and Accuracy.}
  \label{tab:app_ablation1}
  \setlength{\tabcolsep}{2.4mm}
  \renewcommand{\arraystretch}{1.05}
  \resizebox{\textwidth}{!}{%
    \begin{tabular}{cccccccccc}
      \toprule
      SSC & MI & DI & GenEval & Accuracy & PSNR ($\uparrow$) & SSIM ($\uparrow$) & LPIPS ($\downarrow$) & CLIP-IQA ($\uparrow$) & Q-Align ($\uparrow$) \\
      \midrule
      \rowcolor{gray!20}
      & & & 85.01\% & 84.81\% & 16.93 & 0.7943 & 0.2764 & 0.8052 & 0.9404 \\
      \checkmark & & & 86.16\% & 85.76\% & 25.65 & 0.9036 & 0.1041 & 0.7936 & 0.9229 \\
      \checkmark & \checkmark & & 86.06\% & 85.67\% & 25.61 & 0.9035 & 0.1045 & 0.7907 & 0.9210 \\
      \checkmark & & \checkmark & 87.21\% & 86.84\% & 24.78 & 0.8999 & 0.1100 & 0.8105 & 0.9341 \\
      \rowcolor{orange!20}
      \checkmark & \checkmark & \checkmark & 87.32\% & 86.93\% & 24.83 & 0.9049 & 0.1080 & 0.8081 & 0.9315 \\
      \bottomrule
    \end{tabular}%
  }
  \vspace{-4mm}
\end{table*}

\begin{table*}[t!]
  \centering
  \caption{Speed-quality trade-off on BAGEL under varying thresholds. Moderate $\tau_k$ and $\tau_d$ form a stable operating region, while overly relaxed directional constraints lead to fidelity degradation.}
  \label{tab:app_tau-ablation}
  \setlength{\tabcolsep}{2.4mm}
  \renewcommand{\arraystretch}{1.05}
  \resizebox{\textwidth}{!}{%
    \begin{tabular}{cccccccccc}
      \toprule
      $\tau_k$ & $\tau_d$ & Latency (s) & Speedup & GenEval & PSNR ($\uparrow$) & SSIM ($\uparrow$) & LPIPS ($\downarrow$) & CLIP-IQA ($\uparrow$) & Q-Align ($\uparrow$) \\
      \midrule
      \rowcolor{gray!20}
      0.00 & 0.0 & 51.13 & 1.00$\times$ & 85.9149\% & $\infty$ & 1.000000 & 0.000000 & 0.849839 & 0.948901 \\
      0.05 & 0.6 & 13.16 & 3.89$\times$ & 87.0493\% & 24.805368 & 0.899878 & 0.110797 & 0.804107 & 0.931193 \\
      0.05 & 0.7 & 13.54 & 3.78$\times$ & 86.3895\% & 23.691624 & 0.880669 & 0.145908 & 0.787209 & 0.910825 \\
      0.05 & 0.8 & 11.70 & 4.37$\times$ & 83.8022\% & 20.423058 & 0.818049 & 0.255853 & 0.701524 & 0.807565 \\
      0.05 & 0.9 & 10.62 & 4.81$\times$ & 85.1705\% & 19.989533 & 0.807618 & 0.275784 & 0.701656 & 0.841531 \\
      \rowcolor{orange!20}
      0.06 & 0.6 & 12.43 & 4.11$\times$ & 87.3227\% & 24.828357 & 0.904860 & 0.108023 & 0.808102 & 0.931482 \\
      0.07 & 0.7 & 12.54 & 4.08$\times$ & 86.5901\% & 23.196954 & 0.871689 & 0.157621 & 0.789624 & 0.911064 \\
      0.08 & 0.8 & 10.71 & 4.77$\times$ & 85.6350\% & 21.076604 & 0.835902 & 0.215533 & 0.746876 & 0.885157 \\
      \bottomrule
    \end{tabular}%
  }
  \vspace{-4mm}
\end{table*}
\vspace{-2mm}
\section{Pareto Frontier}
\label{app:pareto}
\begin{figure*}[t!]
    \centering
    \begin{subfigure}[t]{0.48\textwidth}
        \centering
        \includegraphics[width=\linewidth]{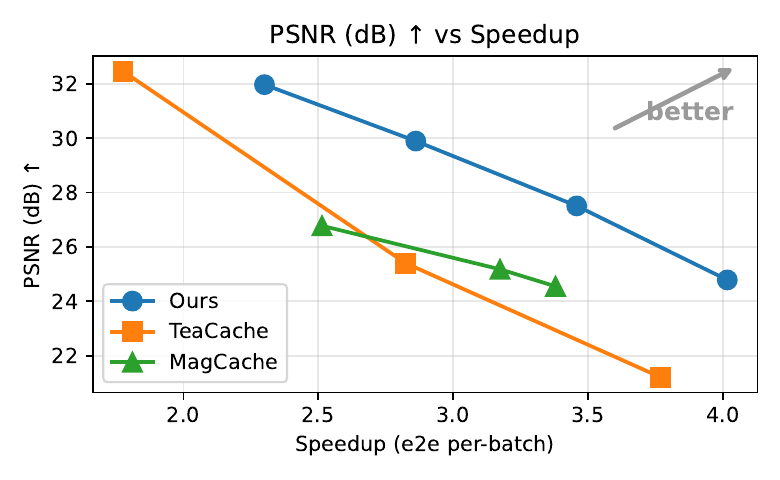}
        \caption{PSNR to the full-step baseline.}
        \label{fig:pareto_psnr}
    \end{subfigure}
    \hfill
    \begin{subfigure}[t]{0.48\textwidth}
        \centering
        \includegraphics[width=\linewidth]{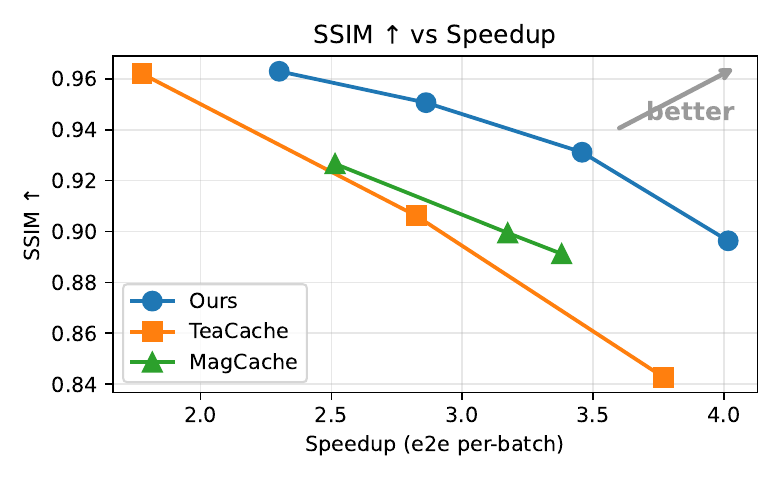}
        \caption{SSIM to the full-step baseline.}
        \label{fig:pareto_ssim}
    \end{subfigure}

    \caption{
    Additional Pareto frontiers on BAGEL. TACache achieves a better
    speed-quality trade-off than TeaCache and MagCache under PSNR and SSIM,
    with a smaller fidelity drop at higher speedups.
    }
    \label{fig:pareto_all}
    \vspace{-0.8em}
\end{figure*}

\vspace{-2mm}
Fig.~\ref{fig:pareto_all} reports additional Pareto frontiers on BAGEL using
reference-based fidelity metrics. TACache maintains a higher speed-quality
frontier than TeaCache and MagCache under both PSNR and SSIM, and the gap
becomes larger as the speedup target increases. In particular, TACache shows a
shallower degradation slope at high speedups, indicating that its cached
updates remain closer to the full-step trajectory under aggressive skipping.
This supports the role of trajectory-aware compensation: restoring both the
parallel magnitude and the orthogonal direction makes the skipped updates more
robust than directly reusing cached features or compensating only the magnitude
component.

\vspace{-2mm}
\section{Detailed Sub-Scores on Main Benchmarks}
\label{app:geneval}
\vspace{-2mm}
We provide detailed sub-score breakdowns for the benchmarks reported in Sec~\ref{sec:exp}. Tables~\ref{tab:gen_eval_bagel} and~\ref{tab:flux_geneval} report GenEval sub-scores on BAGEL~\cite{bagel} and FLUX.1-dev~\cite{flux}, and Table~\ref{tab:wan_vbench} reports VBench sub-scores on Wan2.1-1.3B~\cite{wan}.

% ---------------- BAGEL GenEval ----------------
\begin{table*}[t]
\vspace{-2mm}
  \centering
  \caption{Detailed comparison of GenEval sub-scores (\%) on BAGEL across methods.}
  \setlength{\tabcolsep}{1.4mm}
  \renewcommand{\arraystretch}{1.1}
  \small
  \resizebox{\textwidth}{!}{%
    \begin{tabular}{lccccccccc}
    \toprule
    Method & Overall & Correct Img.\ & Correct Prompt & Position & Color Attr.\ & Counting & Two Obj.\ & Single Obj.\ & Colors \\
    \midrule
    BAGEL (T=50)   & 85.91 & 85.53 & 93.49 & 73.25 & 73.50 & 80.94 & 94.95 & 98.44 & \textbf{94.41} \\
    BAGEL (T=13)   & 85.01 & 84.81 & 94.39 & 74.00 & 74.25 & 74.69 & \textbf{95.71} & 97.81 & 93.62 \\
    TeaCache       & 86.47 & 86.17 & 94.21 & \textbf{75.50} & 76.75 & 80.31 & 94.19 & 98.44 & 93.62 \\
    TaylorSeer     & 82.12 & 81.92 & 94.03 & 73.25 & 67.50 & 67.81 & 94.70 & \textbf{99.06} & 90.43 \\
    A2S            & 87.27 & \textbf{87.03} & \textbf{95.30} & 74.25 & \textbf{82.75} & 81.88 & 95.45 & 97.81 & 91.49 \\
    MagCache       & 87.29 & 86.98 & 95.12 & 75.00 & 78.75 & 81.88 & 95.20 & 98.75 & 94.15 \\
    \rowcolor{orange!20}
    TACache (ours) & \textbf{87.32} & 86.93 & 93.85 & 74.00 & 78.75 & \textbf{84.38} & 94.44 & 98.75 & 93.62 \\
    \bottomrule
    \end{tabular}%
  }
  \label{tab:gen_eval_bagel}%
\end{table*}

Table~\ref{tab:gen_eval_bagel} presents the GenEval sub-score breakdown on BAGEL. TACache attains the highest Overall score of $87.32\%$, with the gain concentrated in counting and color attribute binding. A2S leads on correct images, correct prompts, and color attribute binding; across the other categories the two methods are close, and TACache retains the best overall balance among accelerated methods.

% ---------------- FLUX GenEval ----------------
\begin{table*}[!t]
  \centering
  \caption{Detailed comparison of GenEval sub-scores (\%) on FLUX.1-dev across methods.}
  \setlength{\tabcolsep}{1.4mm}
  \renewcommand{\arraystretch}{1.1}
  \small
  \resizebox{\textwidth}{!}{%
    \begin{tabular}{lccccccccc}
    \toprule
    Method & Overall & Correct Img.\ & Correct Prompt & Position & Color Attr.\ & Counting & Two Obj.\ & Single Obj.\ & Colors \\
    \midrule
    FLUX.1-dev (T=50) & 64.27 & 62.57 & 62.57 & 18.00 & 45.00 & \textbf{71.25} & 73.74 & \textbf{100.00} & 77.66 \\
    FLUX.1-dev (T=13) & 63.87 & 62.21 & 62.21 & \textbf{21.00} & 40.00 & 70.00 & 74.75 & 98.75 & 78.72 \\
    TeaCache          & 62.86 & 61.30 & 61.30 & 15.00 & 40.00 & 63.75 & \textbf{81.82} & \textbf{100.00} & 76.60 \\
    TaylorSeer        & 62.65 & 60.94 & 60.94 & 16.00 & 41.00 & \textbf{71.25} & 77.78 & 97.50 & 72.34 \\
    A2S               & 59.01 & 57.14 & 57.14 & 13.00 & 39.00 & 65.00 & 63.64 & \textbf{100.00} & 73.40 \\
    MagCache          & \textbf{64.61} & \textbf{63.11} & \textbf{63.11} & 18.00 & \textbf{46.00} & 66.25 & 80.81 & \textbf{100.00} & 76.60 \\
    \rowcolor{orange!20}
    TACache (ours)    & 63.71 & 62.03 & 62.03 & 18.00 & 42.00 & 68.75 & 73.74 & \textbf{100.00} & \textbf{79.79} \\
    \bottomrule
    \end{tabular}%
  }
  \label{tab:flux_geneval}%
\end{table*}

Table~\ref{tab:flux_geneval} reports the GenEval sub-score breakdown on FLUX.1-dev. Overall scores on this backbone are limited by the Position metric across all methods, including the baseline, indicating that spatial understanding is a backbone-level bottleneck rather than an artifact of caching. TACache stays close to the step-truncation baseline at $T=13$, while using the same thirteen-step budget, and preserves the Position score of the full-step baseline. MagCache attains a slightly higher Overall score, while TACache obtains the highest Colors score among all methods.

% ---------------- Wan VBench ----------------
\begin{table*}[!t]
\vspace{-2mm}
  \centering
  \caption{Detailed comparison of VBench sub-scores (\%) on Wan2.1-1.3B across methods. We report eight selected dimensions together with the overall total score and end-to-end inference latency. A2S is omitted because it does not support video diffusion models.}
  \setlength{\tabcolsep}{1.0mm}
  \renewcommand{\arraystretch}{1.1}
  \small
  \resizebox{\textwidth}{!}{%
    \begin{tabular}{lcccccccccc}
    \toprule
    Method & Latency (s) & Subj.\ Cons.\ & Dyn.\ Deg.\ & Mot.\ Smooth.\ & Overall Cons.\ & Aesth.\ Qual.\ & Imag.\ Qual.\ & Scene & Bkg.\ Cons.\ & Total \\
    \midrule
    Wan2.1-1.3B (T=50) & 256.78 & 94.96 & \textbf{70.83} & 97.72 & \textbf{23.02} & 64.47 & \textbf{69.83} & 23.18 & 97.14 & 72.59 \\
    TeaCache           & 134.57 & 95.00 & 69.44 & 97.66 & 22.98 & 64.09 & 69.23 & 23.55 & 97.13 & 72.38 \\
    TaylorSeer         & 122.20 & 94.55 & \textbf{70.83} & \textbf{98.00} & 22.82 & 62.73 & 67.39 & 23.39 & 97.00 & 72.03 \\
    MagCache           & 130.65 & 94.98 & 69.44 & 97.67 & 22.92 & 63.67 & 68.90 & \textbf{26.67} & 97.16 & \textbf{72.77} \\
    \rowcolor{orange!20}
    TACache (ours)     & \textbf{121.60} & \textbf{95.04} & \textbf{70.83} & 97.66 & 22.98 & \textbf{64.52} & 69.35 & 23.98 & \textbf{97.23} & 72.64 \\
    \bottomrule
    \end{tabular}%
  }
  \label{tab:wan_vbench}%
  \vspace{-4mm}
\end{table*}

Table~\ref{tab:wan_vbench} reports the VBench sub-score breakdown on Wan2.1-1.3B. TACache matches the baseline on the overall total score under a $2.11\times$ speedup, without any meaningful aggregate-quality drop overall, and achieves the lowest latency among the accelerated methods. At the dimension level, TACache improves over the baseline on subject consistency, aesthetic quality, scene consistency, and background consistency, while motion smoothness, overall consistency, and dynamic degree remain nearly unchanged. Imaging quality drops from $69.83\%$ to $69.35\%$, the only dimension with a noticeable decrease, reflecting a small loss on fine-detail fidelity under aggressive acceleration.
\vspace{-2mm}
\section{Comparison with Concurrent Works}
\label{app:concurrent}
\vspace{-2mm}

\begin{figure}[H]
  \centering
  \captionsetup{skip=4pt}
  \makebox[\linewidth][c]{%
    \resizebox*{!}{0.62\textheight}{%
      \begin{minipage}[b]{\textwidth}
        \centering
        \includegraphics[width=\linewidth]{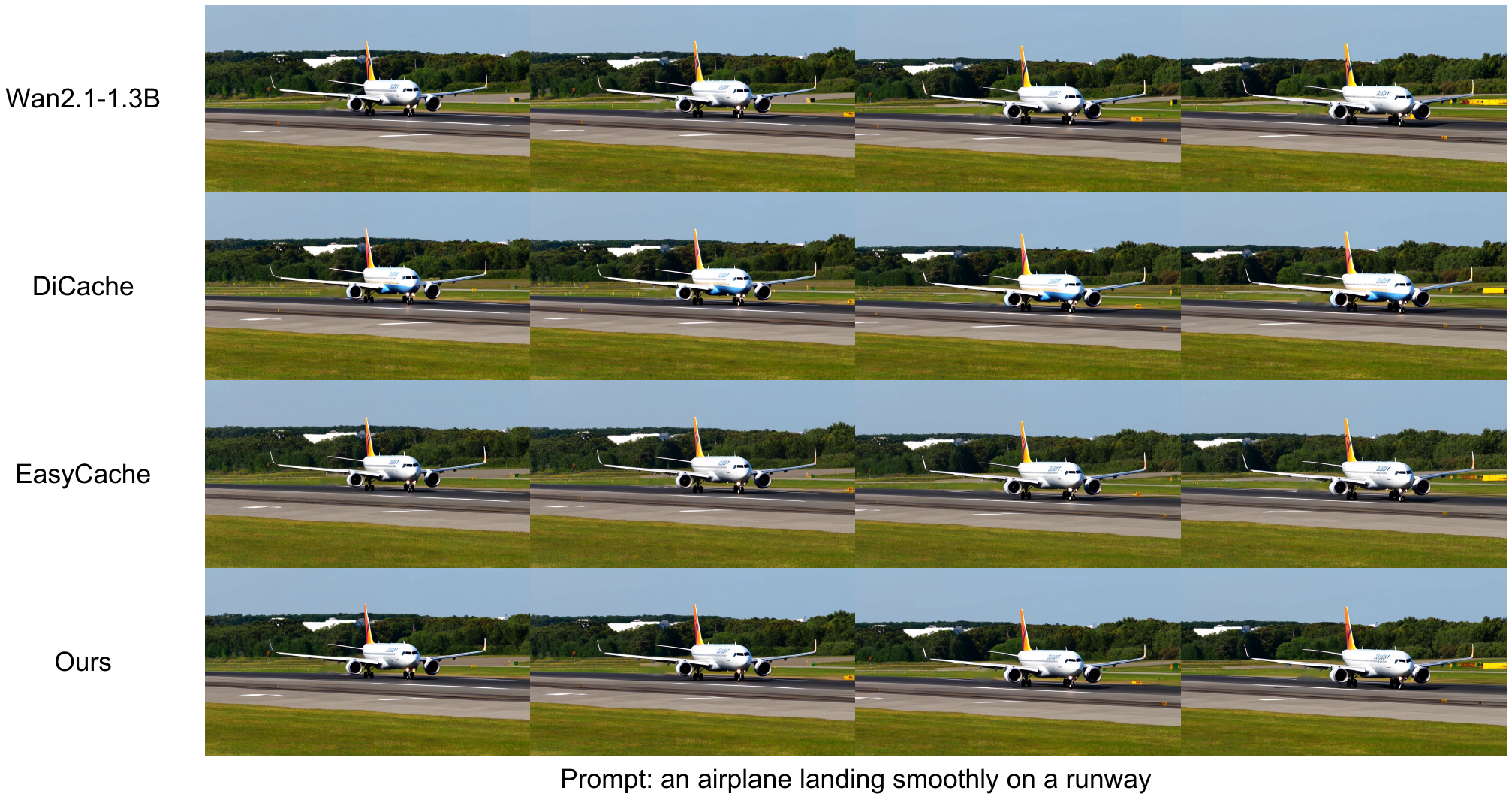}
        \includegraphics[width=\linewidth]{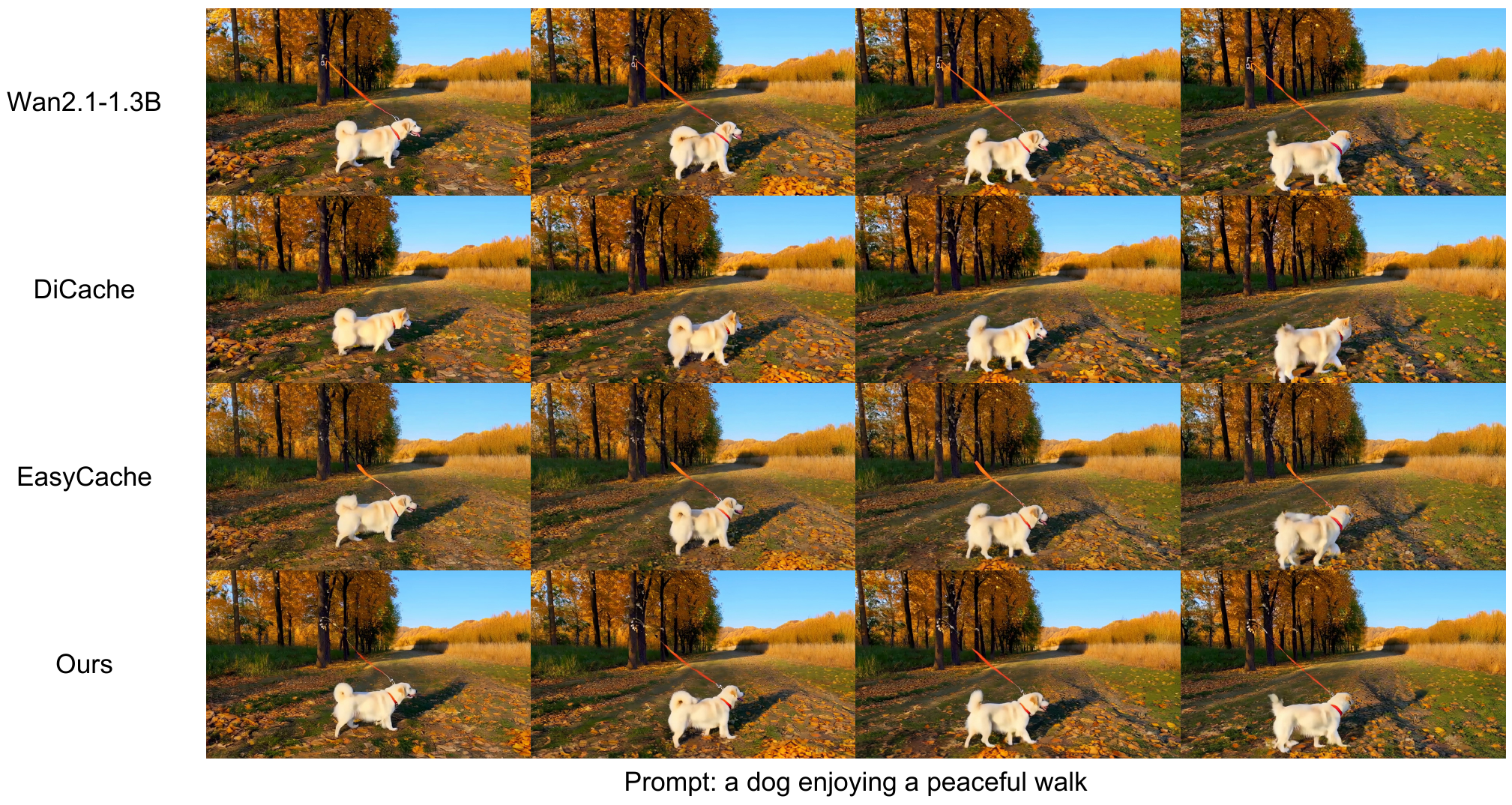}
      \end{minipage}%
    }%
  }
  \caption{Qualitative comparison with concurrent methods on Wan2.1-1.3B.}
  \label{fig:concurrent_wan}
\end{figure}

\begin{figure}[p]
  \centering
  \includegraphics[width=0.95\textwidth]{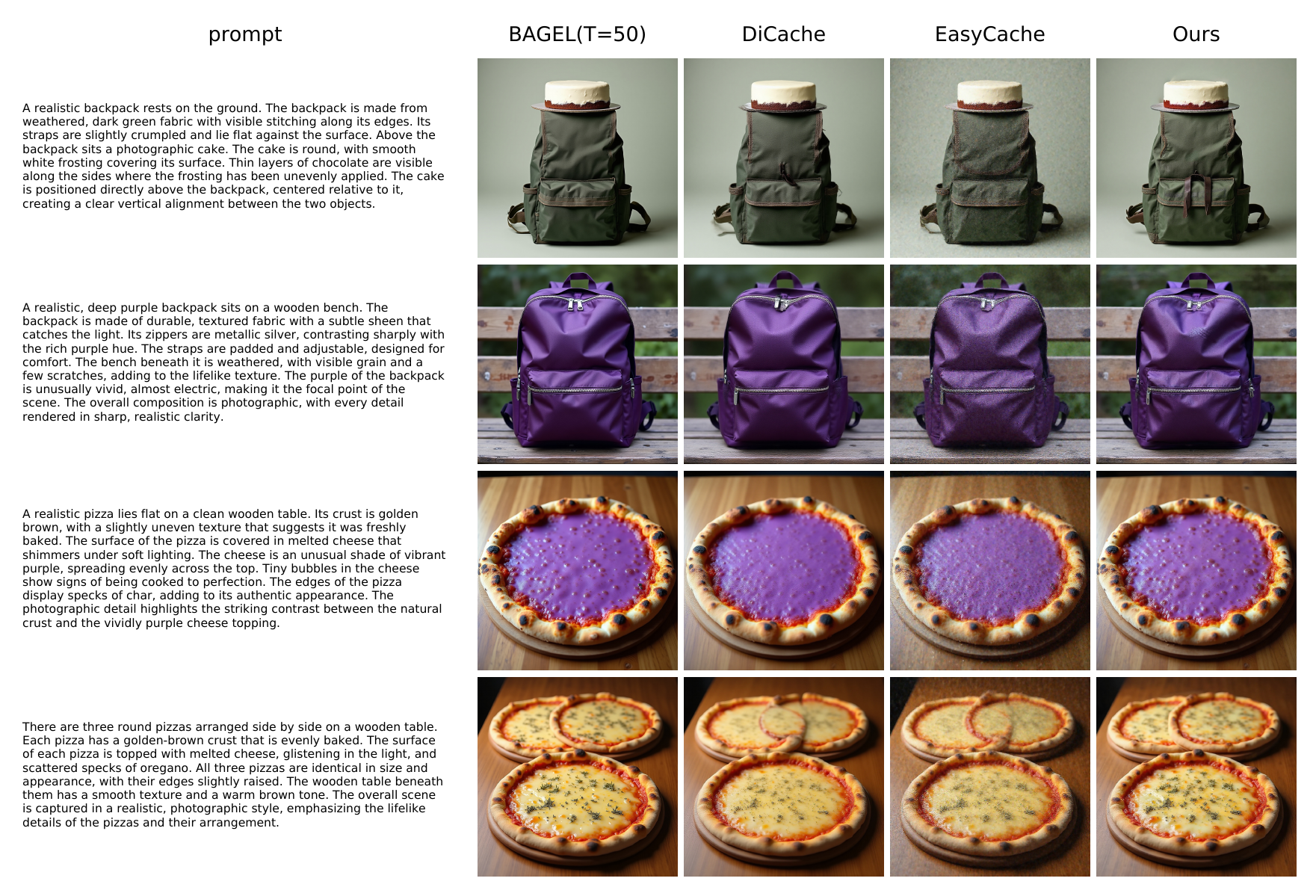}
  \caption{Qualitative comparison with concurrent methods on BAGEL.}
  \label{fig:concurrent_bagel}
  \vspace{-2mm}
\end{figure}

\begin{figure}[p]
  \vspace{-4mm}
  \centering
  \includegraphics[width=0.95\textwidth]{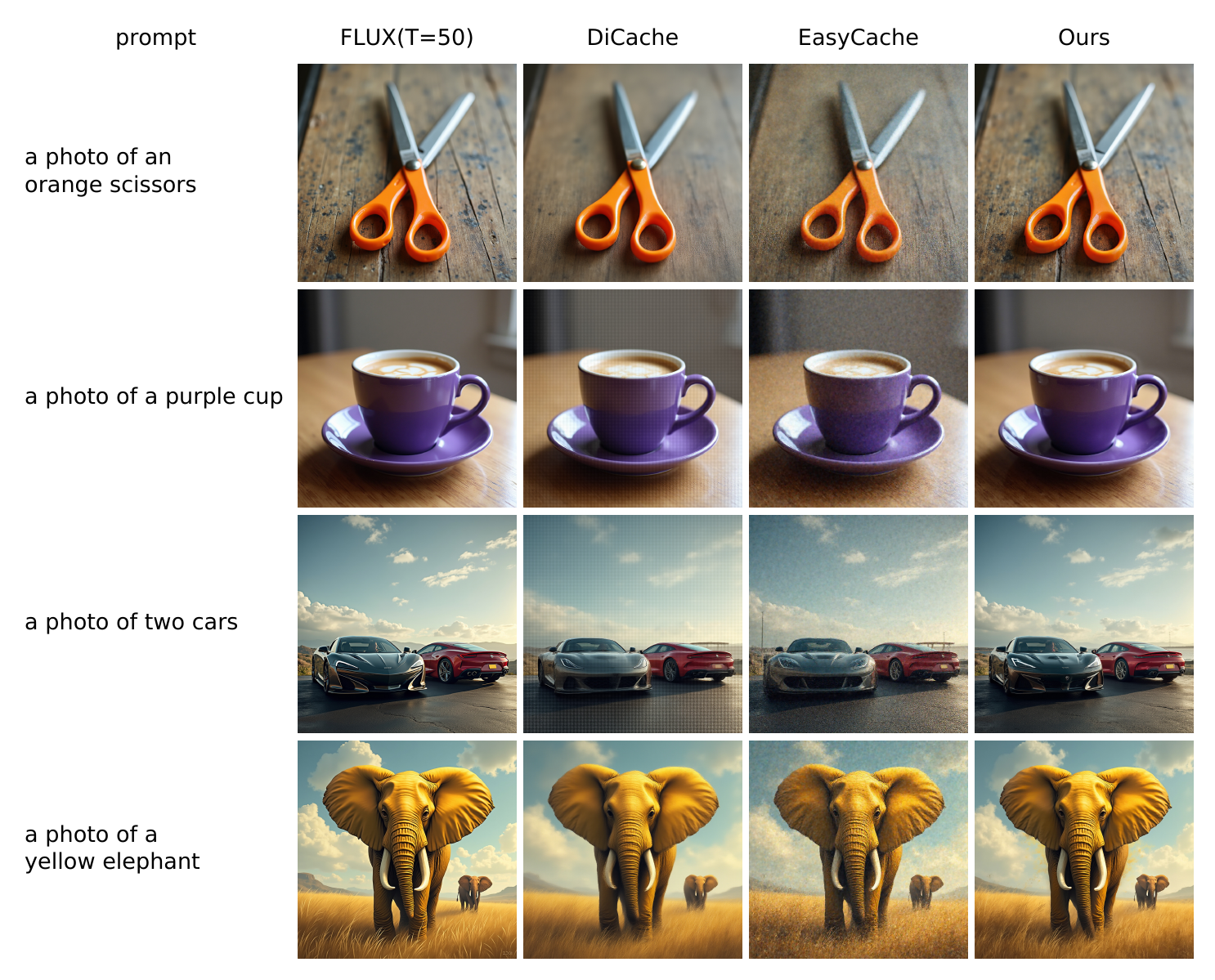}
  \caption{Qualitative comparison with concurrent methods on FLUX.1-dev.}
  \label{fig:concurrent_flux}
\end{figure}

We further compare TACache with concurrent caching methods DiCache~\cite{dicache} and EasyCache~\cite{easycache}.
As shown in Table~\ref{tab:concurrent}, TACache achieves the lowest latency and highest benchmark scores across all three backbones, with the best SSIM and LPIPS and competitive PSNR.
On BAGEL~\cite{bagel}, TACache reduces latency by $27.1\%$ over DiCache while improving GenEval from $85.73\%$ to $87.32\%$.
On FLUX.1-dev~\cite{flux}, it improves GenEval from $61.79\%$ to $63.71\%$, while reducing latency from $10.01$\,s to $8.26$\,s.
On Wan2.1-1.3B~\cite{wan}, TACache achieves the highest VBench score of $72.64\%$ with the lowest latency of $121.60$\,s.
Qualitative comparisons in Figs.~\ref{fig:concurrent_wan}--\ref{fig:concurrent_flux} further demonstrate that TACache preserves visual details closely resembling the full-step outputs.
\vspace{-2mm}
\section{Limitations and Future Work}
\vspace{-2mm}
Our experiments focus on three Rectified Flow backbones, BAGEL, FLUX.1-dev, and Wan2.1-1.3B, covering both text-to-image and text-to-video generation. However, performance on other architectures, longer videos, and higher resolutions remains to be explored.
TACache uses offline trajectory statistics to build a fixed skip schedule for each backbone and sampling configuration. Small calibration sets are effective, but the statistics and schedules may require recalibration for new backbones. Future work may explore more transferable calibration and adaptive schedules.
\clearpage
\begin{table}[H]
  \centering
  \caption{Comparison with concurrent cache-based methods. TACache achieves the lowest latency and the best benchmark and perceptual scores across all three backbones.}
  \label{tab:concurrent}
  \setlength{\tabcolsep}{1.4mm}
  \renewcommand{\arraystretch}{1.1}
  \small
  \resizebox{\textwidth}{!}{%
    \begin{tabular}{lccccccc}
    \toprule
    Method & Latency (s) ($\downarrow$) & Score ($\uparrow$) & PSNR ($\uparrow$) & SSIM ($\uparrow$) & LPIPS ($\downarrow$) & CLIP-IQA ($\uparrow$) & Q-Align ($\uparrow$) \\
    \midrule
    \multicolumn{8}{c}{\textbf{BAGEL (GenEval)}} \\
    \midrule
    DiCache~\cite{dicache}     & 17.0596 & 85.7339\% & \textbf{26.2881} & 0.90353 & 0.11070 & 0.762900 & 0.903871 \\
    EasyCache~\cite{easycache} & 13.17 & 83.73\% & 25.7225 & 0.85415 & 0.31732 & 0.480234 & 0.663827 \\
    \rowcolor{orange!20}
    TACache (ours) & \textbf{12.43} & \textbf{87.32}\% & 24.8300 & \textbf{0.90490} & \textbf{0.10800} & \textbf{0.808100} & \textbf{0.931500} \\
    \midrule
    \multicolumn{8}{c}{\textbf{FLUX.1-dev (GenEval)}} \\
    \midrule
    DiCache~\cite{dicache}     & 10.01 & 61.79\% & 23.1894 & 0.80533 & 0.27606 & 0.785618 & 0.870651 \\
    EasyCache~\cite{easycache} & 8.72 & 57.84\% & \textbf{23.3509} & 0.75782 & 0.41224 & 0.456426 & 0.702337 \\
    \rowcolor{orange!20}
    TACache (ours) & \textbf{8.26} & \textbf{63.71}\% & 22.5700 & \textbf{0.83710} & \textbf{0.19690} & \textbf{0.880800} & \textbf{0.937000} \\
    \midrule
    \multicolumn{8}{c}{\textbf{Wan2.1-1.3B (VBench)}} \\
    \midrule
    DiCache~\cite{dicache}     & 138.49 & 72.42\% & 24.9083 & 0.85816 & 0.07513 & 0.703281 & 0.901568 \\
    EasyCache~\cite{easycache} & 123.89 & 72.50\% & \textbf{26.1792} & 0.84785 & 0.07742 & 0.601085 & 0.901610 \\
    \rowcolor{orange!20}
    TACache (ours) & \textbf{121.60} & \textbf{72.64}\% & 25.1900 & \textbf{0.86370} & \textbf{0.07460} & \textbf{0.709700} & \textbf{0.909600} \\
    \bottomrule
    \end{tabular}%
  }
  \vspace{-5mm}
\end{table}

\clearpage

\clearpage
\section{Additional qualitative results}\label{appfig}

\begin{figure}[H]
    \centering
    \makebox[\linewidth][c]{\includegraphics[height=\dimexpr\textheight-9.5\baselineskip\relax]{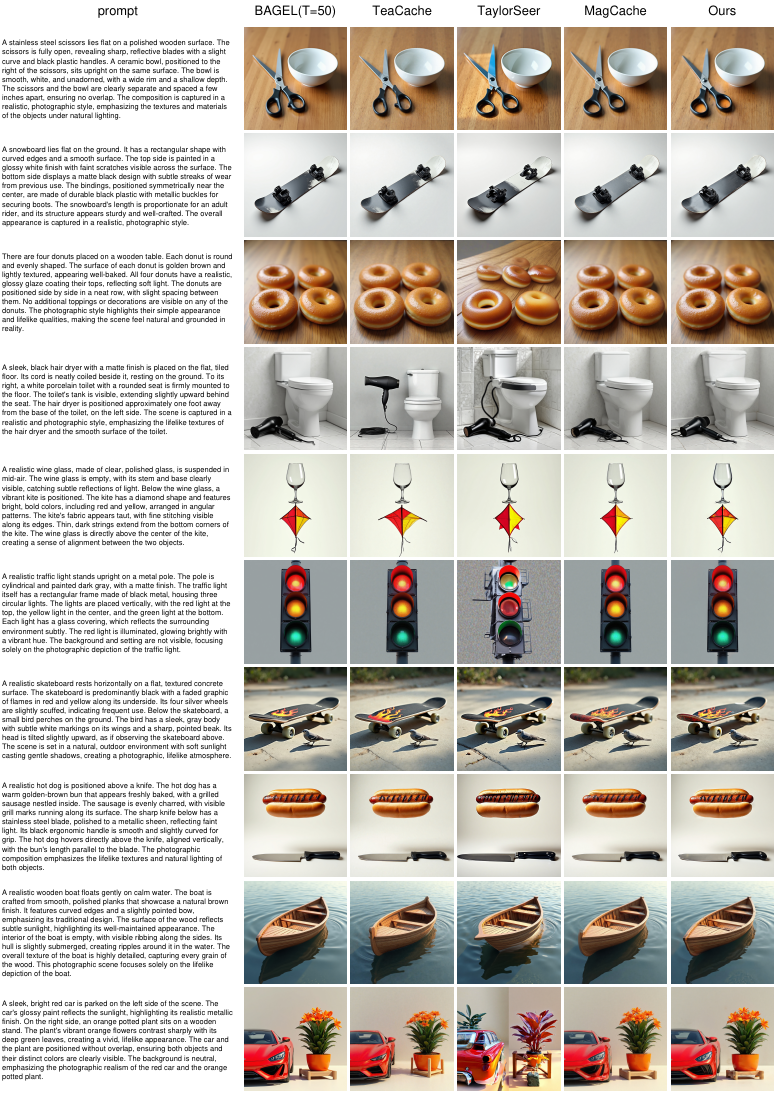}}
    \caption{Visual results of BAGEL}
    \label{fig:appbagel}
\end{figure}

\begin{figure*}[p]
    \centering
    \includegraphics[width=\linewidth]{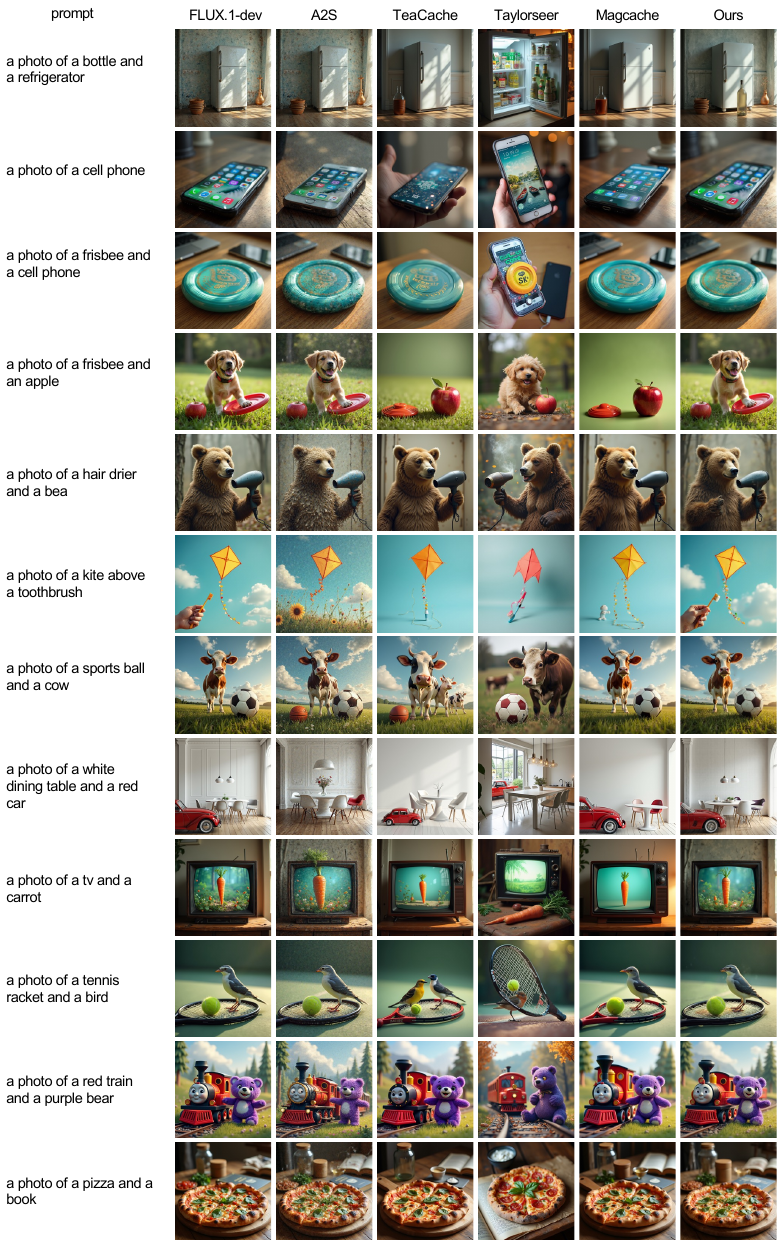}
    \caption{Visual results of FLUX.1-dev}
    \label{fig:appflux}
\end{figure*}

\begin{figure*}[p]
    \centering
    \makebox[\linewidth][c]{\includegraphics[height=\dimexpr\textheight-3em\relax]{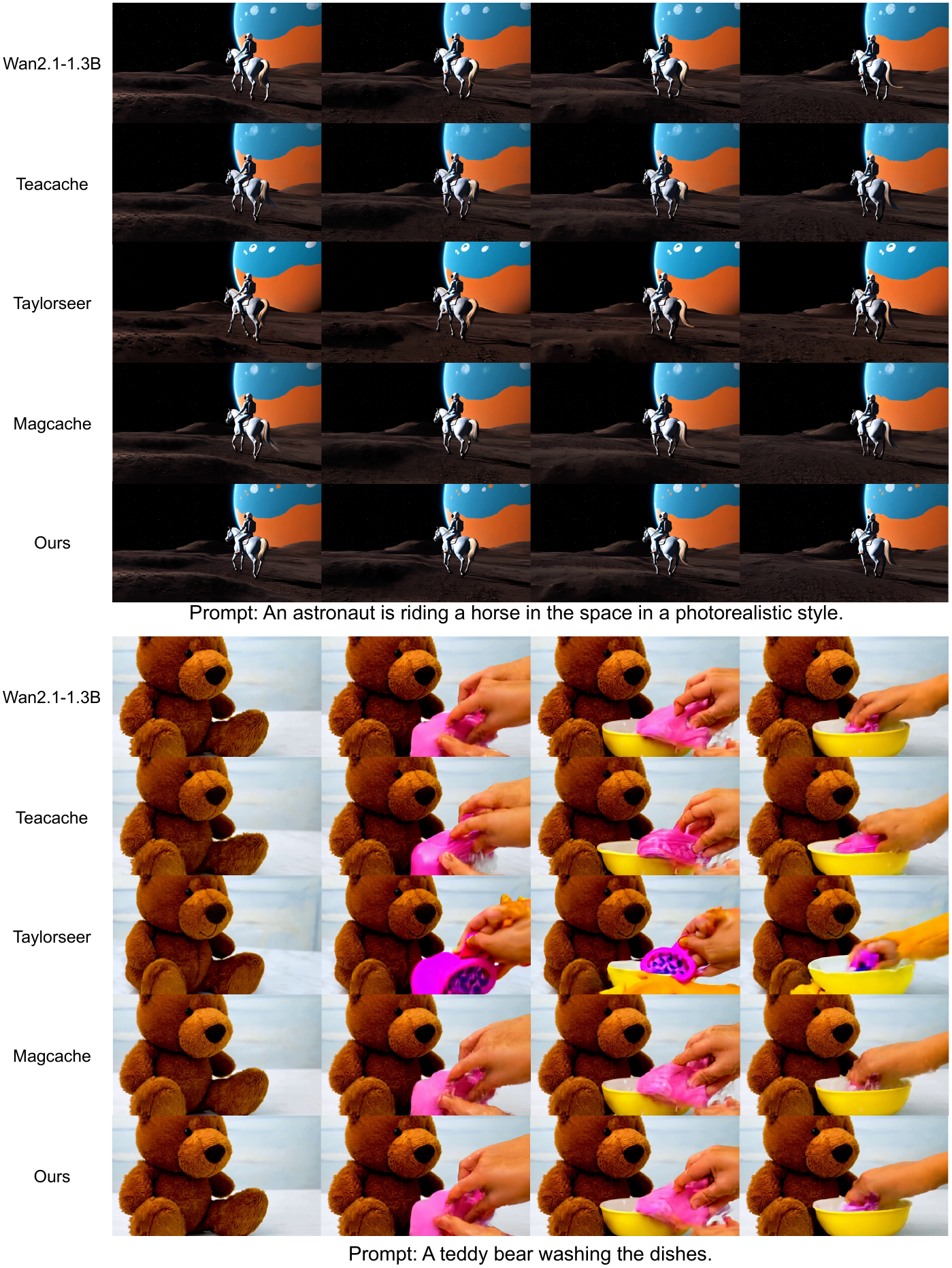}}
    \caption{Visual results of Wan2.1-1.3B}
    \label{fig:appwan1}
\end{figure*}

\begin{figure*}[p]
    \centering
    \makebox[\linewidth][c]{\includegraphics[height=\dimexpr\textheight-3em\relax]{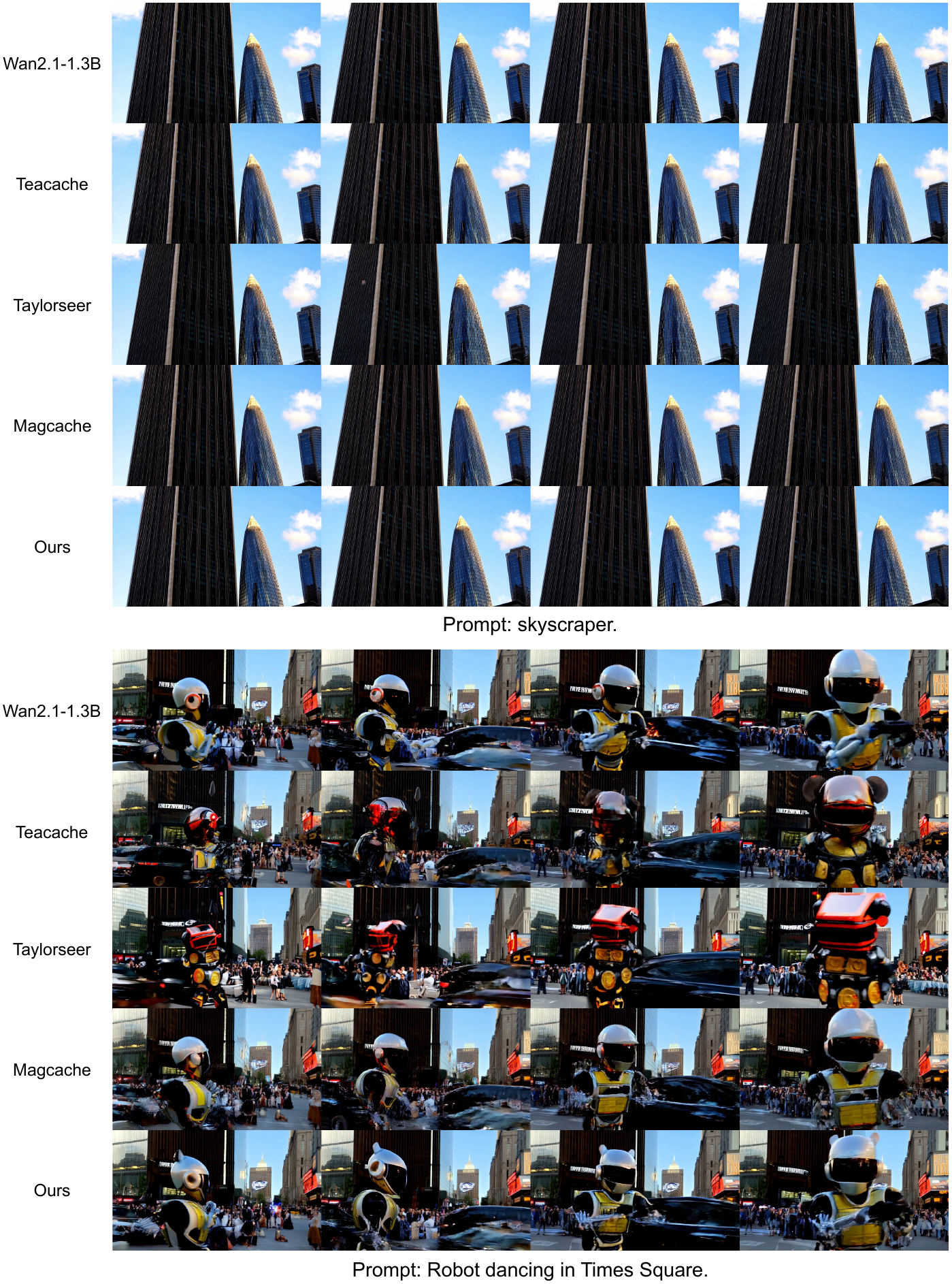}}
    \caption{Visual results of Wan2.1-1.3B}
    \label{fig:appwan2}
\end{figure*}

\begin{figure*}[p]
    \centering
    \makebox[\linewidth][c]{\includegraphics[height=\dimexpr\textheight-3em\relax]{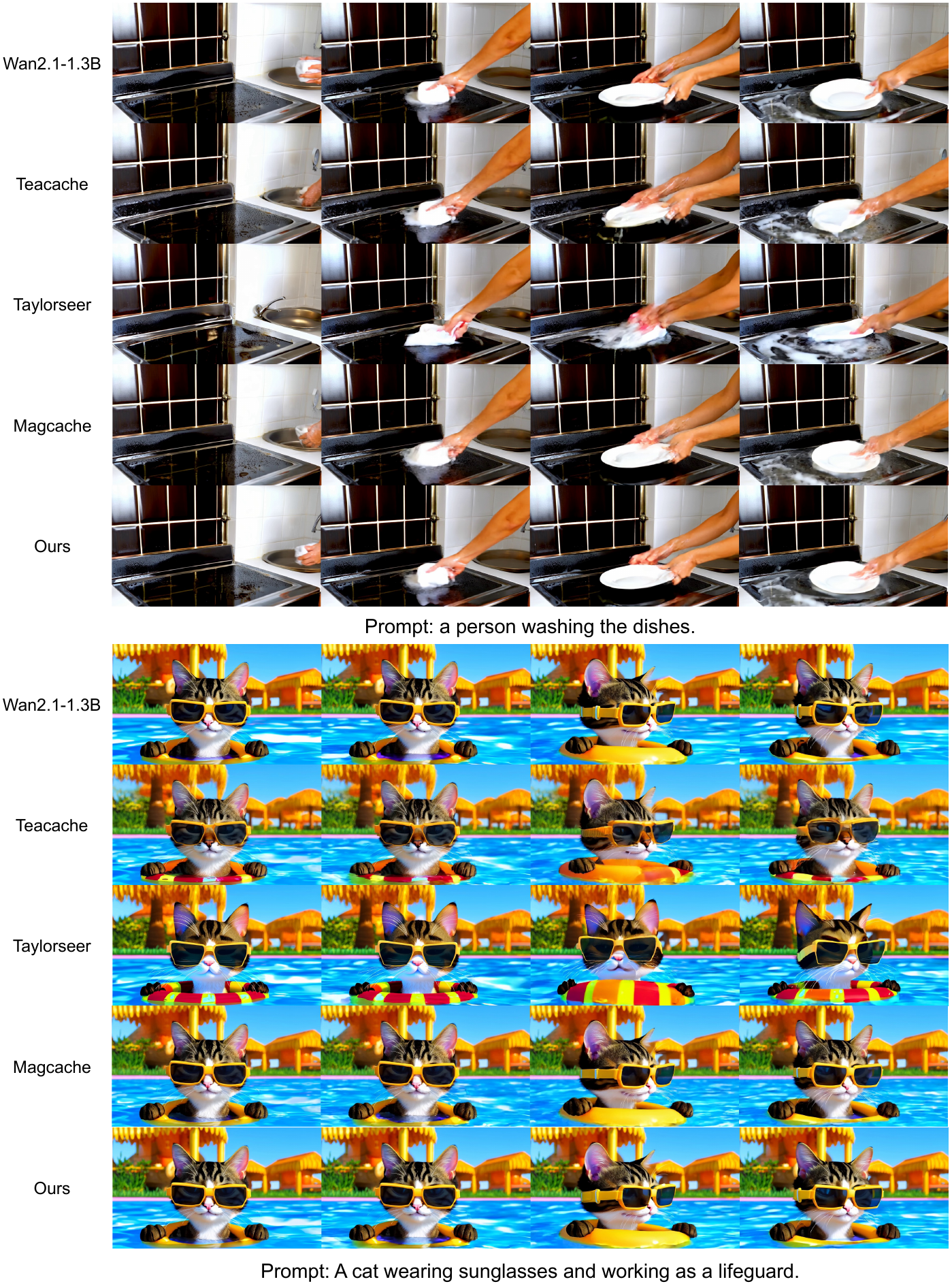}}
    \caption{Visual results of Wan2.1-1.3B}
    \label{fig:appwan3}
\end{figure*}

We present additional results on BAGEL in Fig.~\ref{fig:appbagel}, on FLUX.1-dev in Fig.~\ref{fig:appflux}, and on Wan2.1-1.3B in Figs.~\ref{fig:appwan1}, \ref{fig:appwan2}, and \ref{fig:appwan3}.

\clearpage

\end{document}